\pdfoutput=1
\documentclass{article}

\PassOptionsToPackage{numbers,compress}{natbib}
\usepackage[main,final]{neurips_2026}

\usepackage[utf8]{inputenc}
\usepackage[T1]{fontenc}
\usepackage{url}
\usepackage{booktabs}
\usepackage{amsfonts}
\usepackage{nicefrac}
\usepackage{microtype}
\usepackage[table]{xcolor}
\usepackage{graphicx}
\usepackage{amsmath}
\usepackage{subcaption}
\usepackage{tabularx}
\usepackage{makecell}
\usepackage{sg-ibs-macros}
\usepackage{enumitem}
\usepackage{comment}
\usepackage[pagebackref,breaklinks,colorlinks]{hyperref}
\usepackage{cleveref}

\providecommand{\cmark}{\ensuremath{\checkmark}}
\providecommand{\xmark}{\ensuremath{\times}}

\newcommand{\papertitle}{RoMo Hands: A Large Scale Richly Organized Text to Hand Motion Dataset}
\title{\papertitle}

\author{%
  Yizhak Ben-Shabat$^{1}$ \quad
  Jiahao Zhang$^{1,2}$ \quad
  Joseph Liu$^{1}$ \quad
  Seonghyeon Moon$^{1}$ \\
  Young-Yoon Lee$^{1}$ \quad
  Haomiao Jiang$^{1}$ \quad
  Oren Jacob$^{1}$ \quad
  Mubbasir Kapadia$^{1,3}$ \\[0.5em]
  $^{1}$Roblox \qquad
  $^{2}$Australian National University \qquad
  $^{3}$Rutgers University \\
  \texttt{sitzikbs@gmail.com} \qquad
  \texttt{jiahao.zhang@anu.edu.au}
}

\begin{document}

\maketitle

\begin{center}
  \captionsetup{hypcap=false}
  \includegraphics[width=0.98\linewidth]{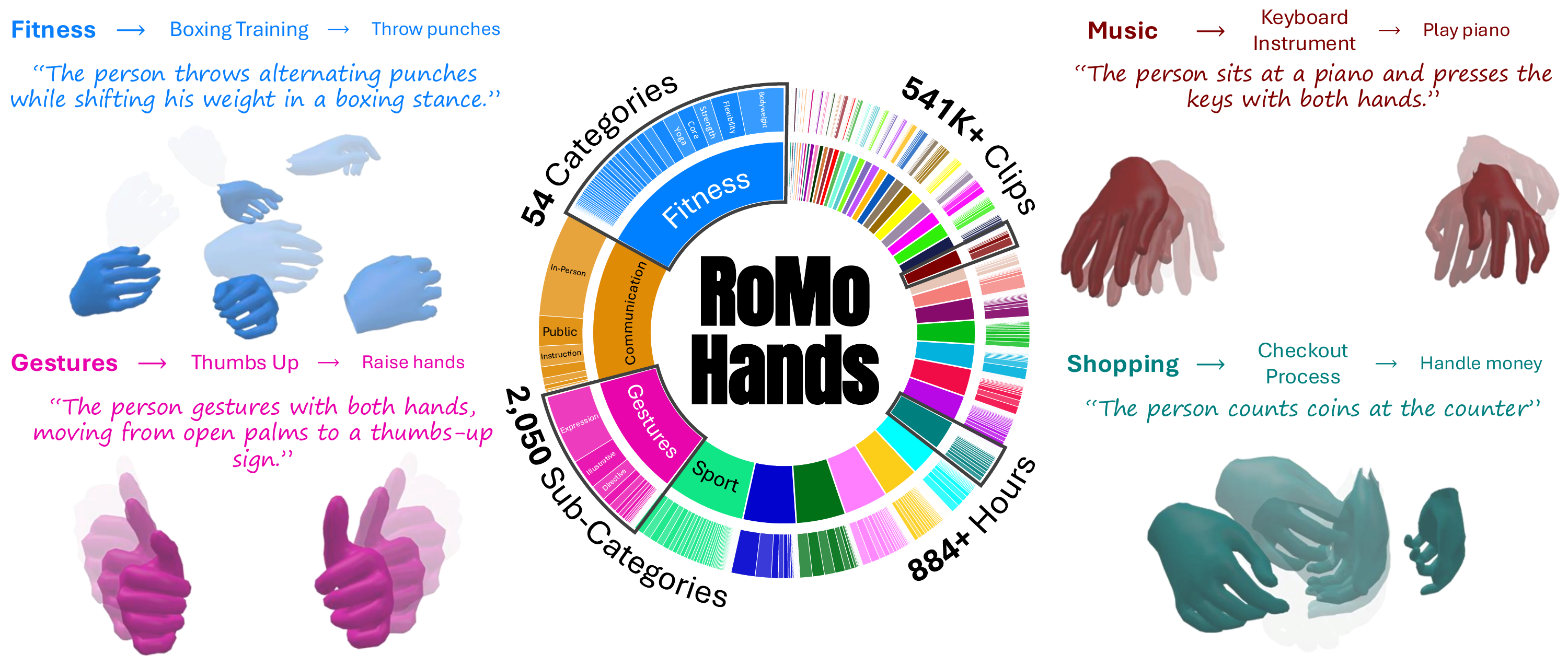}
  \captionof{figure}{\textbf{RoMo Hands scales pure-hand text-to-motion data by an order of magnitude.} $\sim$542K in-the-wild bimanual sequences (884 hours, 95.5M frames at 30\,fps), each paired with five caption granularities (L0--L4) and a three-level action taxonomy. The dataset enables a benchmark of state-of-the-art diffusion, masked-token, and autoregressive generators adapted to a shared 393-D bimanual hand representation. Interactive dataset statistics, side-by-side benchmark comparisons, dataset explorer, and a 3D motion-grid visualizer are best browsed in the supplementary materials.}
  \label{fig:teaser}
\end{center}

\begin{abstract}
While text-driven body motion generation has matured, pure-hand synthesis remains bottlenecked by data. Existing datasets often treat hands as rigid end-effectors, lack fine-grained captions, or are too limited in scale to support robust generalization. We introduce \textbf{RoMo Hands}, an in-the-wild, text-paired dataset of $\sim$542K sequences (884 hours, 95.5M frames at 30\,fps) which is an order of magnitude larger than prior work. To ensure high-quality supervision, a stringent dexterity filter discards low-articulation sequences, focusing the distribution on complex bimanual movement. Each clip is paired with five captions at increasing granularities from global semantic tags to atomic finger states, organized under a hierarchical taxonomy. To bridge the gap between body and hand architectures, we propose a novel 393-D bimanual representation that adapts redundant body-motion paradigms to the high-DoF kinematics of hands. Systematic benchmarking of diffusion, masked-token, and autoregressive generators reveals distinct trade-offs under these hand-specific adaptations, including a sharp divergence between semantic alignment and kinematic plausibility. Our dataset, representation, and benchmark establish a rigorous foundation for the next generation of dexterous motion synthesis. Project page: \url{https://romodataset.github.io/romo-hands/}.
\end{abstract}

\section{Introduction}
\label{sec:intro}

Text-driven human motion synthesis has matured rapidly in recent years, driven by large-scale datasets paired with natural-language captions that enable models to learn rich semantic correspondences between text and motion~\citep{humanml3d,motionx,motionlib,motionmillion}. These advances have substantially improved the realism, controllability, and generalization of body-motion generation systems. \emph{Pure hand} motion, however, has not progressed at the same pace. Yet, hands are the primary channel through which humans manipulate the world and convey nuanced intent. High-fidelity finger articulation is therefore essential for applications such as AR/VR avatars, teleoperation, sign-language synthesis, embodied agents, and dexterous robotics. From a kinematic perspective, hands are uniquely challenging: they pack extremely high degrees of freedom into a compact spatial region and exhibit frequent self-contact together with tightly coordinated inter-finger motion. Existing body-centric generators were not designed for this precision-intensive regime, and the lack of sufficiently large text-paired hand datasets has limited the community’s ability to systematically adapt modern generative architectures to dexterous hand motion.

\tabref{tab:comprehensive_dataset_comparison} and \figref{fig:dataset-comparison} illustrate the sharp trade-off between scale and hand-motion fidelity in current datasets. Large-scale body-centric datasets~\citep{motionx,interact,motionlib,motionmillion,romo_body} provide extensive motion diversity and textual supervision, but typically treat hands as rigid end-effectors with generic captions. In contrast, hand-object interaction and egocentric datasets~\citep{gigahands,hot3d,arctic,h2o,oakink2} capture accurate contact dynamics and realistic manipulation behavior, yet rarely exceed a few hours or include free-form language annotations. Existing pure-hand datasets~\citep{interhand,handdiffuse} similarly lack text supervision, limiting their usefulness for text-conditioned generation. More recent text-paired efforts such as BOTH2Hands~\citep{both2hands}, HandX~\citep{handx}, and the concurrent CLUTCH~\citep{clutch} begin to address this gap, but remain orders of magnitude smaller than modern body-motion corpora. Compounding this challenge, hand-motion generation demands substantially higher fidelity than full-body synthesis: subtle artifacts such as inter-finger penetration or implausible coordination are immediately perceptible, yet are rarely captured by metrics designed for coarse body dynamics.

To address this bottleneck, we introduce \textbf{RoMo Hands}: a pure-hand, text-paired dataset containing approximately 542K in-the-wild motion sequences, corresponding to 884 hours and 95.5M frames at 30\,fps, making it more than an order of magnitude larger than the next-largest pure-hand dataset. To ensure high-quality supervision, we design a stringent automated filtering pipeline that removes near-static and low-information sequences, focusing the retained distribution on genuinely dexterous hand activity. Each clip is paired with five fine-grained captions at increasing levels of detail, describing atomic finger states, temporal dynamics, and spatial relations within a hierarchical action taxonomy.

Beyond the dataset itself, we establish a comprehensive benchmark for pure-hand motion generation. Because leading generative architectures were originally developed for coarse-grained body motion, directly applying them to hands introduces substantial representational challenges. To bridge this gap, we propose a novel redundant bimanual representation, $\mathcal{R}_{\text{hand}}$, specifically designed for dexterous hand synthesis. Using this representation, we systematically adapt diffusion, masked-token, and autoregressive generation paradigms within a unified benchmark setting. Under these evaluated adaptations, semantic and physical metrics yield different model rankings: strong text alignment does not guarantee kinematic plausibility or fine-grained physical consistency. These findings suggest that pure-hand generation introduces challenges beyond those typically encountered in conventional body-motion synthesis. Our core contributions are:

\begin{itemize}[leftmargin=1.4em,topsep=2pt,itemsep=1pt]
    \item \textbf{RoMo Hands Dataset.} A large-scale, in-the-wild pure-hand motion dataset, an order of magnitude beyond existing pure-hand sets.
    \item \textbf{Fine-Grained Captions.} Five hierarchical captions per clip detailing finger articulation, temporal dynamics, and spatial relations, indexed under a pruned RoMo~\citep{romo_body} action taxonomy for stratified analysis.
    \item \textbf{Bimanual Motion Representation.} A 393-dimensional hand representation ($\mathcal{R}_{\text{hand}}$) that adapts the redundant body-motion paradigm to hands via a synthetic root and inter-wrist heading.
    \item \textbf{Pure-Hand Benchmark.} A comprehensive evaluation of diffusion, masked-token, and autoregressive architectures under documented hand-specific adaptations, exposing a tension between semantic and physical quality.
\end{itemize}

\begin{table*}[t]
    \centering
    \caption{\textbf{RoMo Hands provides an order-of-magnitude leap in hand-motion scale.} It surpasses prior pure-hand datasets by $10\times$ in duration and text volume.}
    \label{tab:comprehensive_dataset_comparison}
    \resizebox{\linewidth}{!}{
    \begin{tabular}{
    l l l r r c c 
    }
    \toprule
    \textbf{Dataset} & \textbf{Scope} & \textbf{Setting} & \textbf{Duration (h)} & \textbf{Frames (M)} & \textbf{Granularity} & \textbf{\#Captions (K)} \\
    \midrule
    GRAB \citep{grab}              & Hands + Object & Studio       & 3.8   & 1.6  & ---          & ---    \\
    OakInk2 \citep{oakink2}        & Hands + Object & Studio       & 37.1  & 4.01 & Action       & ---    \\
    ARCTIC \citep{arctic}          & Hands + Object & Studio       & 2.02  & 0.22 & Action       & ---    \\
    H2O \citep{h2o}                & Hands + Object & Studio       & 1.06  & 0.11 & Action       & ---    \\
    HOT3D \citep{hot3d}            & Hands + Object & Studio       & 3.90  & 0.42 & ---          & ---    \\
    GigaHands \citep{gigahands}    & Hands + Object & Egocentric   & 34.0  & 3.7  & Coarse       & 84.0   \\
    HoloAssist \citep{holoassist}  & Hands + Object & Egocentric   & 161.2 & 17.4 & Coarse       & 1.8    \\
    \midrule
    HandDiffuse \citep{handdiffuse}      & Pure Hands     & Studio       & 2.0   & 0.25 & ---          & ---    \\
    BOTH2Hands \citep{both2hands}        & Pure Hands     & Studio       & 8.31  & 1.8  & Coarse       & 23.5   \\
    InterHand2.6M \citep{interhand}      & Pure Hands     & Studio       & 24.0  & 2.6  & ---          & ---    \\
    HandX \citep{handx}                  & Pure Hands     & Studio       & 54.2  & 5.9  & Fine-grained & 485.7  \\
    3D-HIW \citep{clutch}                & Pure Hands     & In-the-wild  & 83.3  & 12.0 & Fine-grained & 32.0   \\
    \midrule
    \rowcolor{blue!5}
    \textbf{RoMo Hands (Ours)}           & \textbf{Pure Hands} & \textbf{In-the-wild} & \textbf{884} & \textbf{95.5} & Fine-grained & \textbf{2710} \\
    \bottomrule
    \end{tabular}
    }
    \end{table*}

\section{Related Work}
\label{sec:related_work}

We categorize prior work along three axes: (1) body-motion scaling paradigms; (2) hand-centric datasets and their scale constraints; and (3) generative architectures adapted to the bimanual domain.

\subsection{Large-Scale Body Motion Datasets}
\label{sec:related_body}
Text-to-motion research has transitioned from studio captures like AMASS~\citep{amass} and HumanML3D~\citep{humanml3d} to massive, in-the-wild datasets such as Motion-X~\citep{motionx}, MotionLib~\citep{motionlib}, and MotionMillion~\citep{motionmillion}. While these efforts scale body motion by orders of magnitude, they prioritize gross locomotion and treat hands as rigid end-effectors~\citep{motionx, motionmillion, romo_body}. RoMo Hands imports the successful scaling recipe of RoMo~\citep{romo_body}, leveraging in-the-wild video and automated filtering to introduce the first body-scale, text-paired dataset devoted strictly to dexterous articulation.

\subsection{Hand and Hand-Object Interaction Datasets}
\label{sec:related_hand_data}
A disjoint line of work focuses on hands (\tabref{tab:comprehensive_dataset_comparison}). Studio mocap datasets like InterHand2.6M~\citep{interhand}, GRAB~\citep{grab}, ARCTIC~\citep{arctic}, and OakInk2~\citep{oakink2} provide precise MANO~\citep{mano} parameters but span only a few hours and rarely include free-form text. Egocentric captures like GigaHands~\citep{gigahands} and HoloAssist~\citep{holoassist} broaden behavioral coverage but produce coarse annotations targeted at assistive agents rather than high-fidelity synthesis.

Pure-hand text-paired datasets are nascent but have made crucial strides. BOTH2Hands~\citep{both2hands} records 8.31 hours of motion, while HandX~\citep{handx} unifies datasets via LLM-based annotation (54.2 hours). Concurrent to our work, CLUTCH~\citep{clutch} introduces an in-the-wild pure-hand dataset of $\sim$83 hours. These efforts establish the foundation for text-to-hand generation, yet they remain far below the scale required for the data-hungry architectures dominating the body domain. RoMo Hands provides this missing scaling engine with 884 hours, $\sim$542K sequences, and dense five-level captioning.

\subsection{Text-Conditioned Motion Generation}
\label{sec:related_models}
Generation has converged on three paradigms. \emph{Diffusion} models (e.g., MotionDiffuse~\citep{motiondiffuse}, MDM~\citep{mdm}, PriorMDM~\citep{priormdm}, Kimodo~\citep{kimodo}) iteratively denoise continuous sequences and lead the high-fidelity regime. \emph{Autoregressive} models (e.g., T2M-GPT~\citep{t2m_gpt}, MotionMillion~\citep{motionmillion}) predict discrete tokens over VQ-VAE or FSQ~\citep{fsq} codebooks. Finally, \emph{masked-token} models like MoMask~\citep{momask} and SnapMoGen~\citep{snapmogen} utilize bidirectional transformers for joint token generation.

Multiple prior works target text-to-HOI~\citep{text2hoi,diffh2o,gears}, however, they rely on fixed object assets. For pure hands, CLUTCH~\citep{clutch} represents a recent end-to-end system. However, it remains unclear how other generative architectures behave on the unique, high-DoF kinematics of hands. Rather than reinventing architectures, we ask how the existing models can be adapted. By bridging them with our novel representation ($\mathcal{R}_{\text{hand}}$) and benchmarking on RoMo Hands, we isolate which body-centric design choices transfer to dexterous articulation and which require non-trivial adaptation.

\section{The RoMo Hands Dataset and Representation}
\label{sec:approach}

\begin{figure}[t]
  \centering
  \includegraphics[width=0.98\linewidth]{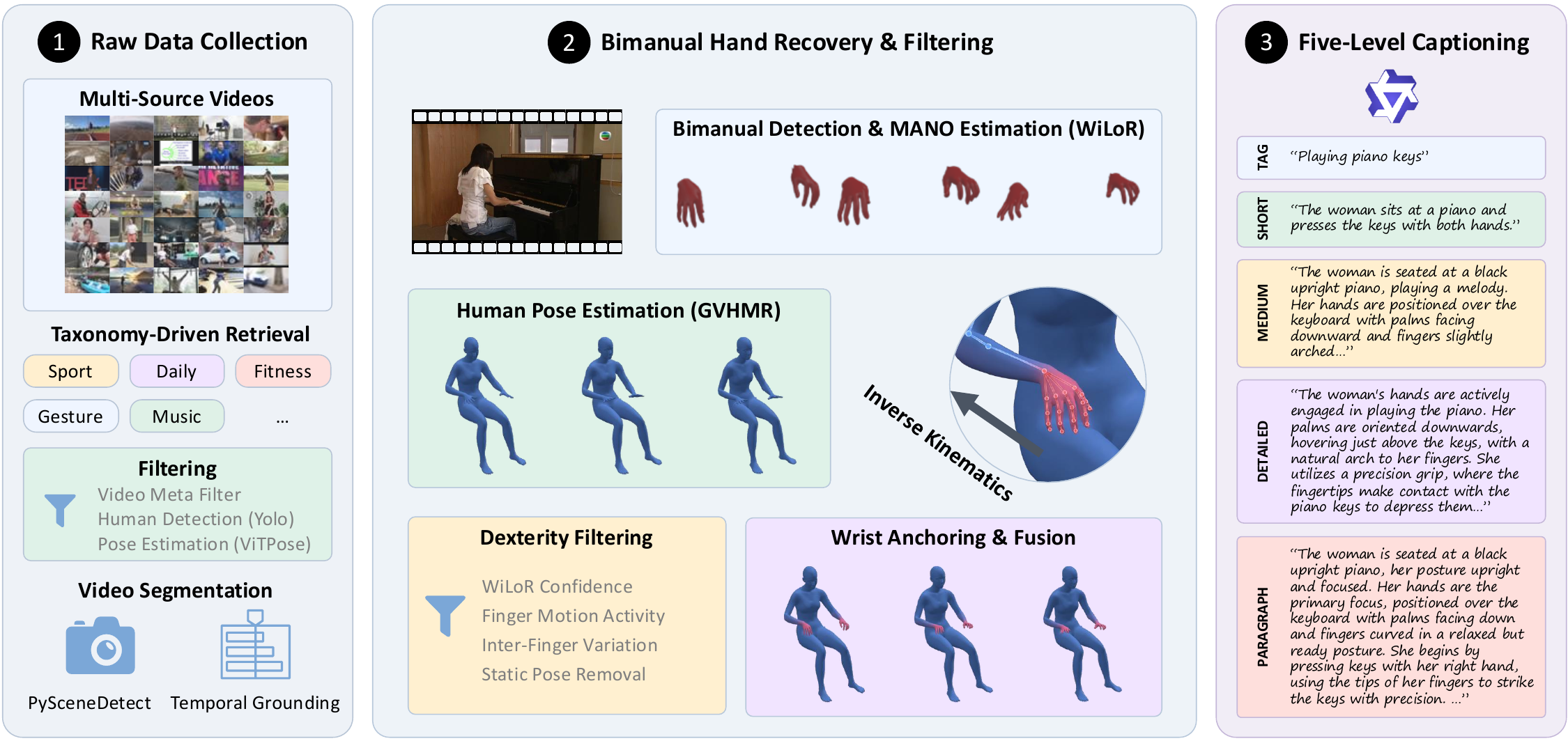}
  \caption{\textbf{The RoMo Hands curation pipeline.} It extends the RoMo body-motion stack with three hand-specific stages: hand detection and bimanual MANO pose recovery (WiLoR), a dexterity filter (detection-ratio + finger-articulation thresholds), and a five-level VLM captioner (Qwen3.5).}
  \label{fig:collection-pipeline}
\end{figure}

\subsection{Collection and Captioning}
\label{sec:collection}

\paragraph{The RoMo Hands collection pipeline.}
RoMo Hands is built upon the data-collection and curation pipeline introduced by RoMo~\citep{romo_body}, which we rigorously repurpose for fine-grained pure-hand motion. The RoMo pipeline begins with a multi-source web-video collection assembled via taxonomy-driven queries. They apply metadata filtering, scene detection, and single-human filtering, followed by semantic temporal segmentation against a three-level action taxonomy. They then estimate 3D body motion in SMPL format for every retained clip using GVHMR~\citep{gvhmr}, caption each segment using a vision-language model, and finally prune the dataset with a category-adaptive dynamic-score filter that removes static or low-activity sequences.

Inheriting RoMo's video pool, we extend this pipeline along three axes specifically targeted at hand motion failure modes. First, we introduce a bimanual hand-pose stage that recovers per-frame MANO parameters using WiLoR~\citep{wilor} and reconciles them against the inherited wrist trajectory. Second, we enforce a strict visibility threshold, discarding sequences where the mean per-frame WiLoR detection confidence falls below a required per-hand minimum, ensuring the retained clips contain consistently visible hands rather than heavily occluded or off-screen artifacts. Third, we utilize Qwen3.5-27B~\citep{qwen3.5} to caption each segment with five text descriptions at increasing levels of specificity, prompting the LLM to emphasize object manipulation context, temporal hand dynamics, and inter-finger spatial relationships. \figref{fig:collection-pipeline} summarizes the end-to-end stack.

\paragraph{Bimanual hand recovery.}
We recover hand motion using WiLoR~\citep{wilor}, a robust monocular bimanual hand estimator that performs per-frame hand detection and MANO~\citep{mano} parameter regression in a single pass. Each frame yields up to two hand instances, carrying a left/right label, axis-angle wrist orientation, 15 finger-joint rotations, weak-perspective camera parameters $\{\mathbf{t}_{cam}, \mathbf{s}_{cam}\}$, and a detection score cached for downstream filtering.

Because WiLoR lacks a temporal component and assumes a weak-perspective camera, per-frame wrist position and orientation suffer from frame-by-frame ambiguity: absolute depth is entangled with hand scale, and wrist orientation lacks a shared anchor across frames. We resolve both issues via a per-frame \emph{wrist anchoring} step that reconciles WiLoR's wrist pose against the temporally consistent SMPL~\citep{smpl} wrist pose recovered by RoMo's body pipeline. When the two estimates agree within a configurable threshold, we average them. Otherwise, we trust WiLoR's wrist pose and solve inverse kinematics (IK) on the body's arm chain to bring the SMPL wrist into agreement, locking both hands into the body's temporally consistent world frame. For isolated frames where WiLoR fails to detect a hand, we linearly interpolate the MANO parameters in axis-angle space between the nearest detections. This interpolation is reliable only for short gaps (up to 5 frames); longer gaps trigger a segment split or discard.
In aggregate, the hand-specific pipeline (bimanual recovery, plus the visibility and dexterity filters described next) retains $\sim$884\,h of the $\sim$1{,}238\,h inherited from RoMo~\citep{romo_body}, a $\sim$29\% reduction.

\paragraph{Detection-ratio and dexterity filtering.}
Even after segmentation, two failure modes dominate the raw distribution: (a) clips where hands are barely visible due to occlusion or motion blur, and (b) clips where hands are visible but remain near a static rest pose (e.g., resting at the side or statically holding an object). Training a generator on these modes degrades quality by injecting regression noise or biasing the model toward the mean hand pose.

We address this with two complementary filters. First, a \emph{detection-ratio filter} discards clips if the fraction of high-confidence WiLoR frames falls below a per-hand minimum. Second, a \emph{dexterity filter} discards clips failing a threshold that combines finger-articulation activity (mean joint-rotation velocity over the 15 finger joints) and inter-finger pose variation (mean pairwise $\ell_2$ distance between fingertip positions over time). Crucially, both filters run \emph{before} captioning to preserve VLM budget and bias the final distribution toward dexterous, well-observed motion.

\paragraph{Captioning at five granularities.}
Downstream tasks consume hand motion at varying detail levels as retrieval benefits from short headlines, and generation from paragraph-length descriptions. We therefore caption each clip at five progressive granularities via a single structured prompt to Qwen3.5-27B~\citep{qwen3.5}, letting downstream models select supervision level at training time and enabling retrieval evaluation as a function of caption complexity:
\begin{itemize}[leftmargin=1.4em,topsep=2pt,itemsep=1pt]
\item \textbf{L0 (headline)}: A 2--4 word headline naming the action and salient object, e.g., ``hold volleyball''.
\item \textbf{L1 (one-sentence scenario)}: A short sentence placing the action in body and scene context, e.g., ``The woman holds a volleyball in her left hand while standing on a court.''
\item \textbf{L2 (multi-sentence description)}: A balanced description covering the active hand pose and the role of the passive hand.
\item \textbf{L3 (half-paragraph description)}: Adds finger-level articulation (palm orientation, finger curl) and brief temporal dynamics.
\item \textbf{L4 (full-paragraph description)}: The most verbose level, adding environmental context, fine wrist-and-finger detail, and inferred intent.
\end{itemize}

\paragraph{Human validation.}
We audit caption fidelity and recovered-pose plausibility on 250 randomly sampled validation clips (50 per caption level). Twelve adult raters (ages 20--50) with general technical backgrounds (at least a bachelor's degree), most without animation experience, viewed each source video beside its 3D reconstruction and assigned five-point Likert scores for overall action match, hand-specific caption--motion faithfulness, and reconstruction quality. Participation was uncompensated. The audit collected only technical judgments of the displayed artifacts and no personal or sensitive data; it did not constitute human-subjects research. Across 370 ratings, action match averages $4.16/5$ ($81\%$ at least 4), caption--motion faithfulness averages $3.79/5$ ($66\%$ at least 4), and reconstruction quality averages $3.53/5$. The per-level breakdown in \tabref{tab:user_study} shows that hand-specific caption accuracy remains within $3.70$--$3.96$ across the five levels.

The resulting vocabulary is analyzed in \secref{sec:dataset_stats} and visualized in the right panel of \figref{fig:dataset-scale-vocab}.

\paragraph{Action taxonomy.}
We adopt RoMo's three-level coarse-to-fine taxonomy (\emph{category} $\to$ \emph{sub-category} $\to$ \emph{atomic action})~\citep{romo_body} and prune categories emptied by our visibility and dexterity filters; no new categories are introduced. We use the pruned taxonomy as a stratification axis for evaluation, surfacing per-category model behavior that aggregate metrics hide (\Cref{fig:taxonomy_radar}).

\subsection{Dataset Statistics and Coverage}
\label{sec:dataset_stats}

\begin{figure}[t]
    \centering
    \begin{minipage}{0.95\linewidth}
        \centering
        \begin{subfigure}[b]{0.62\linewidth}
            \centering
            \includegraphics[width=\linewidth]{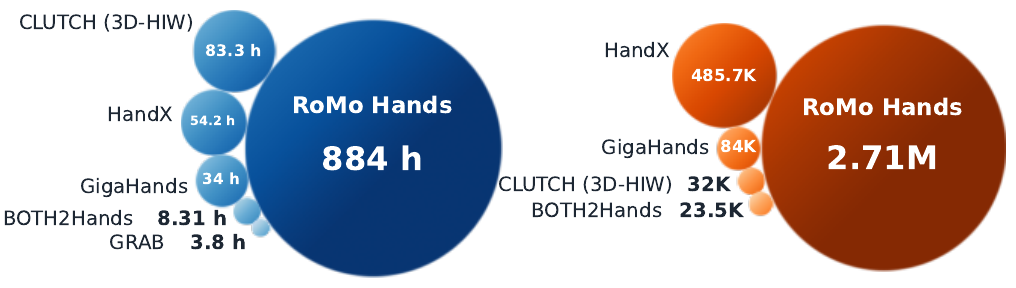}
            \subcaption{Dataset scale: duration and caption counts.}
            \label{fig:pure_hands_circles}
        \end{subfigure}\hfill
        \begin{subfigure}[b]{0.36\linewidth}
            \centering
            \includegraphics[width=\linewidth]{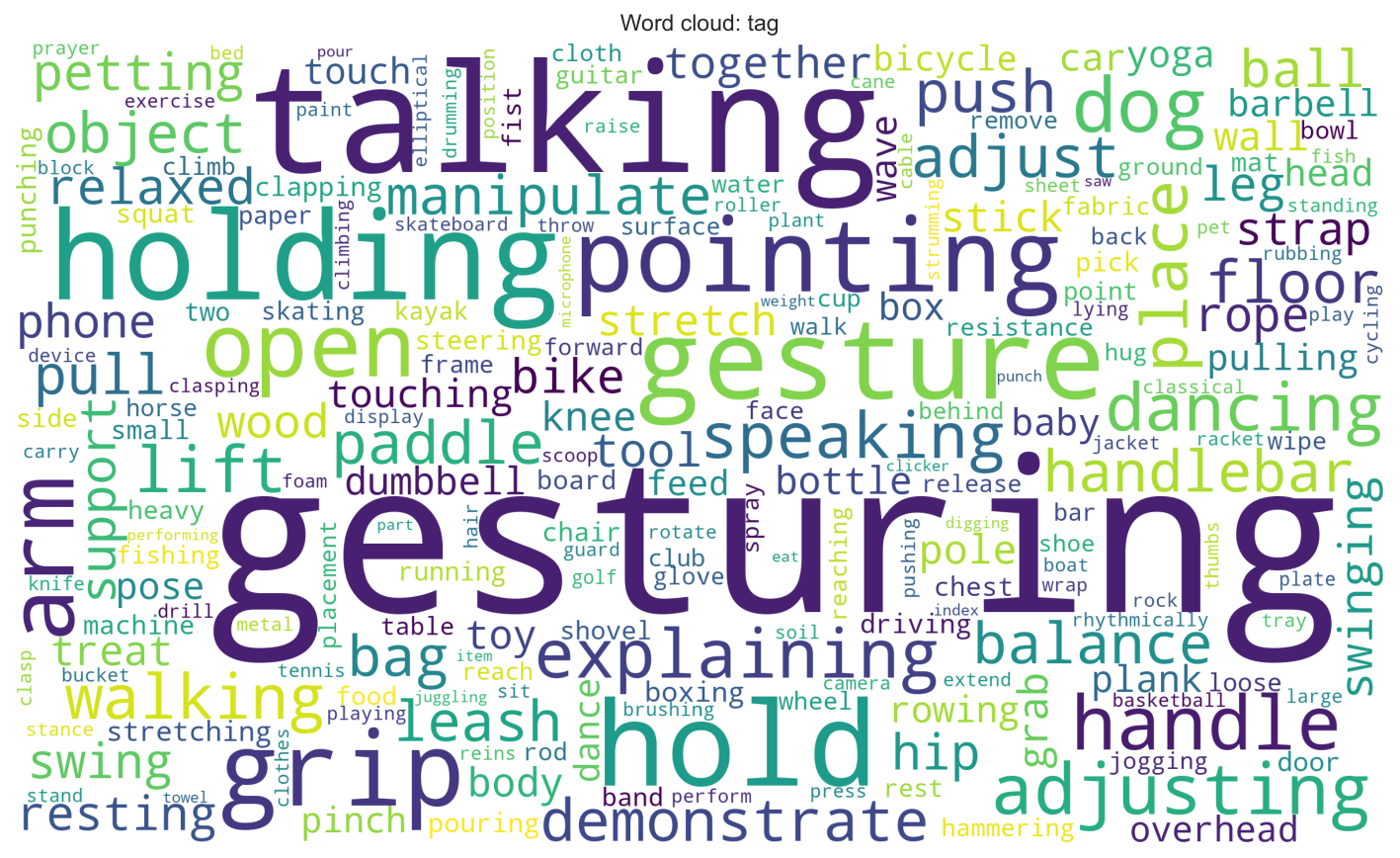}
            \subcaption{Caption-token word cloud.}
            \label{fig:wordcloud_tag}
        \end{subfigure}
    \end{minipage}
    \caption{\textbf{RoMo Hands surpasses prior hand-motion datasets in both scale and caption specificity.} (left)~bubble areas track duration and caption count, with RoMo Hands an order of magnitude beyond the largest prior pure-hand dataset on both axes; GRAB is text-free and omitted from the right bubble. (right)~caption-token word cloud, dominated by finger-level verbs and concrete interaction targets rather than body-motion verbs.}
    \label{fig:dataset-scale-vocab}
\end{figure}

RoMo Hands totals $\sim$542K bimanual hand-motion sequences (884 hours, 95.5M frames at 30\,fps), with a mean clip duration of 5.9\,s and an 85\%/10\%/5\% train/val/test split at the segment level. The left panel of \figref{fig:dataset-scale-vocab} situates this scale against prior pure-hand and hand--object datasets, placing RoMo Hands at $10.6\times$ the duration of CLUTCH and $5.6\times$ the caption count of HandX. To anchor activity coverage, we score GRAB~\citep{grab} ($1.1$K seq., $2.86$\,h) and GigaHands~\citep{gigahands} ($12.7$K seq., $16.40$\,h) under the same dynamic-score function used during curation: RoMo Hands lands at $0.2\%$ static / $53.7\%$ active, versus $20.4\%/14.2\%$ for GigaHands and $0.0\%/65.6\%$ for GRAB, with a mean dynamic score of $0.33$ that sits between GRAB's choreographed $0.66$ and GigaHands' uncurated $0.14$.

\begin{figure*}[t]
  \centering
  \begin{subfigure}[b]{0.500\linewidth}
    \centering
    \includegraphics[width=\linewidth]{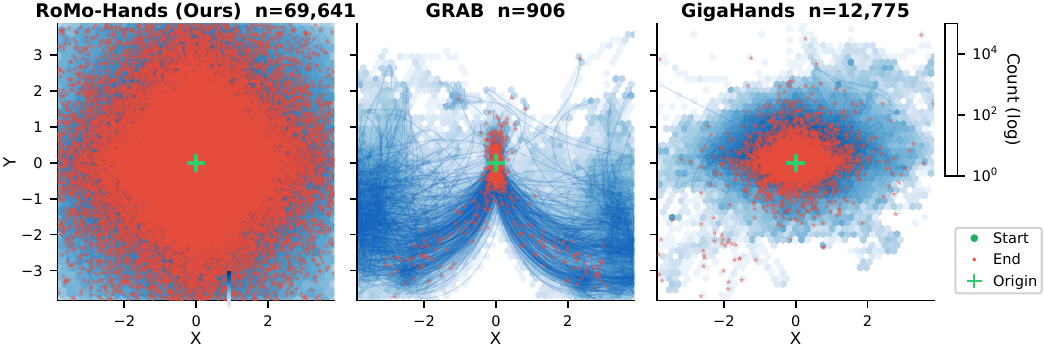}
    \subcaption{Wrist log-density.}
    \label{fig:density-combined}
  \end{subfigure}\hfill
  \begin{subfigure}[b]{0.483\linewidth}
    \centering
    \includegraphics[width=\linewidth]{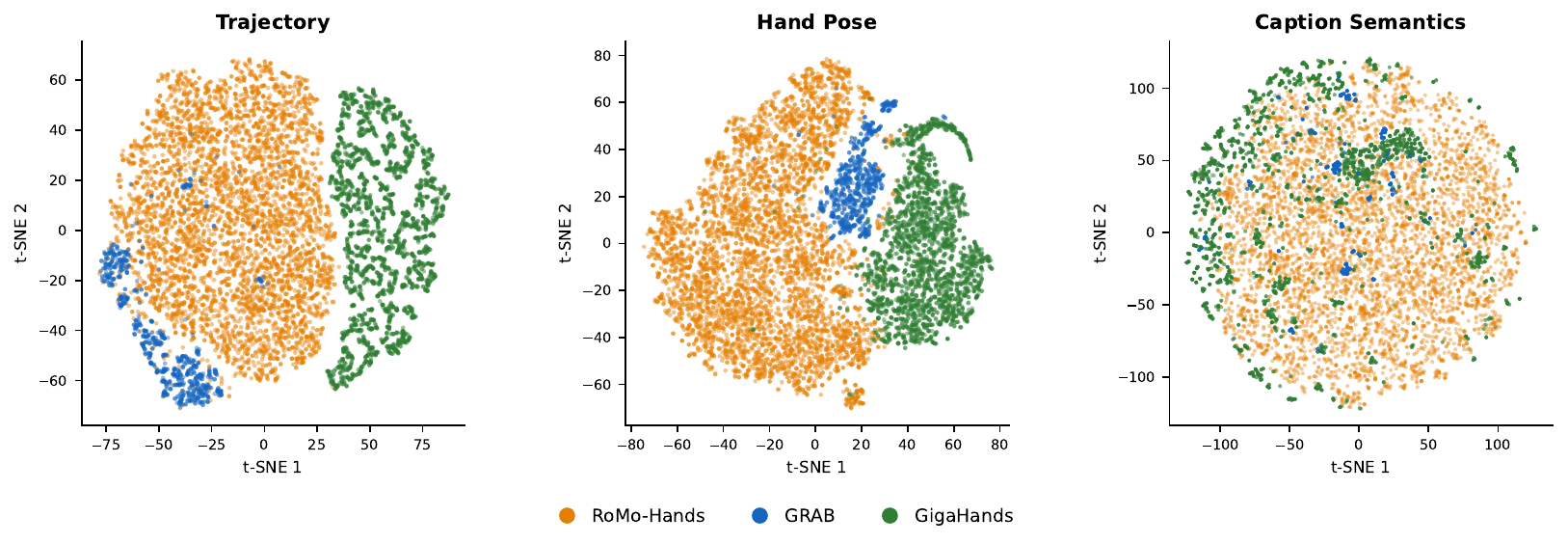}
    \subcaption{t-SNE feature views.}
    \label{fig:tsne-combined}
  \end{subfigure}
  \caption{\textbf{RoMo Hands provides broader spatial, kinematic, and linguistic coverage than prior bimanual datasets.} (a)~Wrist log-density on the workspace plane. (b)~t-SNE in trajectory, hand-pose, and caption-feature spaces, where RoMo Hands spans a wider manifold than GRAB and GigaHands.}
  \label{fig:dataset-comparison}
\end{figure*}

\figref{fig:dataset-comparison} further characterizes spatial, kinematic, and linguistic coverage: wrist XY log-density (left) shows RoMo Hands filling a substantially larger workspace, and t-SNE embeddings in three feature spaces (wrist-trajectory, hand-pose, caption-semantic) reveal a wider manifold along all three axes. The token-frequency view in \figref{fig:dataset-scale-vocab} (right) is dominated by finger-level verbs (\emph{grip}, \emph{pinch}, \emph{flex}) and concrete interaction targets (\emph{phone}, \emph{cup}) rather than body-motion verbs. Per-finger and per-joint dynamic scores, smoothness, autocorrelation, per-tier caption statistics, and side-by-side breakdowns against GRAB and GigaHands are available in the supplementary materials.

\subsection{Bimanual Hand Motion Representation}
\label{sec:hand_repr}

A useful motion representation must faithfully encode skeletal kinematics while plugging cleanly into generative architectures. For body motion, standard templates utilize redundant joint-aware per-frame channels (position, velocity, and rotation) to improve the training signal across diffusion and autoregressive models. The 263-D HumanML3D representation~\citep{humanml3d} and the 272-D representation introduced by MotionStreamer~\citep{motionstreamer} (subsequently adopted by MotionMillion~\citep{motionmillion}, with public reference implementation by~\citet{lixingxiao272repr}) rely on a low-dimensional heading frame followed by per-joint blocks. However, no analogous template has emerged for bimanual hand motion. Existing hand generators like CLUTCH~\citep{clutch} train on a $198$-dimensional MANO-derived representation that lacks the redundant velocity and heading-canonicalized channels that make body templates effective.

We propose $\mathcal{R}_{\text{hand}} \in \mathbb{R}^{393}$, which extends MANO into a redundant per-frame template. We wrap the parametric skeleton in explicit position, velocity, and heading-canonicalized rotation channels, specialized to MANO's 32-joint bimanual topology (16 joints per hand). $\mathcal{R}_{\text{hand}}$ stacks four per-frame channel groups:
\begin{equation}
  \mathbf{h}_t =
  \bigl[
      \underbrace{
        \vphantom{\{\dot{\tilde{\mathbf{p}}}^j_t\}_{j=0}^{31}}
        \mathbf{f}_t}_{9, \text{ root frame}},
      \underbrace{\{\tilde{\mathbf{p}}^j_t\}_{j=0}^{31}}_{96, \text{ joint positions}},
      \underbrace{\{\dot{\tilde{\mathbf{p}}}^j_t\}_{j=0}^{31}}_{96, \text{ joint velocities}},
      \underbrace{\{\phi(\mathbf{R}^j_t)\}_{j=0}^{31}}_{192, \text{ 6D rotations}}
  \bigr]
  \in \mathbb{R}^{393},
  \label{eq:hand_repr_template}
\end{equation}
where $\mathbf{f}_t$ is the 9D root frame defined below, $\tilde{\mathbf{p}}^j_t$ are joint positions whose $xz$ components are root-relative in the heading frame while the $y$ component is the absolute world height, $\dot{\tilde{\mathbf{p}}}^j_t$ are their first-order finite differences, and $\phi: SO(3) \to \mathbb{R}^6$ encodes joint rotations via 6D continuous parameterization~\citep{zhou2019continuity}. The 32 joints follow MANO's bimanual indexing: joints 0--15 and 16--31 for the left and right hands respectively, with wrists at $j=0$ and $j=16$.

Because the bimanual hand has no skeletal joint at its kinematic center, $\mathcal{R}_{\text{hand}}$ introduces two bimanual-specific design choices: \textbf{(i)~a synthetic root} which is the wrist midpoint $\mathbf{r}_t = \tfrac{1}{2}(\mathbf{p}^L_t + \mathbf{p}^R_t)$, stored explicitly in $\mathbf{f}_t$ as 2 channels of planar velocity in the heading frame, 1 channel of world-$y$ height, and 6 channels of heading rotation, with the two actual wrist joints ($j \in \{0, 16\}$) retained in the per-joint blocks alongside the fingers; and \textbf{(ii)~an inter-wrist heading} which is a yaw rotation $\mathbf{R}_y(\psi_t)$ about the world-up axis that aligns the inter-wrist segment with the world $x$-axis (where $\psi_t = \operatorname{atan2}(-(\mathbf{p}^R_t-\mathbf{p}^L_t)_z, (\mathbf{p}^R_t-\mathbf{p}^L_t)_x)$), applied to per-joint $xz$ positions and the two wrist rotations only, while finger rotations remain parent-relative within each MANO chain. Restricting canonicalization to world-up yaw preserves absolute world height, which carries critical semantic context for two-handed gestures.

Because several channels in $\mathcal{R}_{\text{hand}}$ have near-zero variance, we replace standard per-channel z-score normalization with a type-aware scheme (detailed in the supplementary material). Ultimately, this four-block layout, synthetic root, and normalization scheme yield a robust bimanual representation that slots natively into modern text-to-motion architectures.

\begin{table}[t]
\centering
\caption{\textbf{The redundant $\mathcal{R}_{\text{hand}}$ representation improves text alignment across two generators.} Each block holds the architecture, data, optimizer, training budget, text encoder, evaluator, and normalization fixed while changing only the motion representation. Best within each model block is \textbf{bold}.}
\label{tab:representation_ablation}
\small
\setlength{\tabcolsep}{5pt}
\begin{tabular}{@{}llrrrr@{}}
\toprule
\textbf{Generator} & \textbf{Representation}
  & \textbf{R@1} $\uparrow$
  & \textbf{R@3} $\uparrow$
  & \textbf{FID} $\downarrow$
  & \textbf{MM-Dist} $\downarrow$ \\
\midrule
MoMask & $\mathcal{R}_{\text{hand}}$-393 & \textbf{0.547} & \textbf{0.794} & \textbf{0.153} & \textbf{0.941} \\
       & MANO-102                         & 0.395          & 0.658          & 0.221          & 1.049 \\
       & CLUTCH-198                       & 0.368          & 0.617          & 0.269          & 1.077 \\
\addlinespace[3pt]
Kimodo M & $\mathcal{R}_{\text{hand}}$-393 & \textbf{0.665} & \textbf{0.871} & 0.114          & \textbf{0.868} \\
         & MANO-102                         & 0.646          & 0.868          & 0.135          & 0.896 \\
         & CLUTCH-198                       & 0.646          & 0.856          & \textbf{0.106} & 0.882 \\
\bottomrule
\end{tabular}
\end{table}

\paragraph{Representation ablation.}
\tabref{tab:representation_ablation} isolates the representation while holding each generator's data, architecture, optimization, text encoder, and frozen evaluator fixed. For MoMask, $\mathcal{R}_{\text{hand}}$ improves every reported metric over raw MANO and CLUTCH. Kimodo M is less representation-sensitive: $\mathcal{R}_{\text{hand}}$ leads R@1, R@3, MM-Dist, and CLIP, while CLUTCH has the best FID. Thus the redundant position, velocity, and heading channels provide a useful generation-side inductive bias across two distinct architectures, although the magnitude and metric-wise effect are architecture dependent. Within each generator all variants share the same data, architecture, and training budget, so differences reflect the representation; the Kimodo~M models were retrained for this ablation and are not directly comparable to \tabref{tab:results_main}. Full results are in \tabref{tab:representation_ablation_full}.

\section{Experiments}
\label{sec:results}

The goal of this benchmark is to evaluate how state-of-the-art text-to-motion architectures that were originally designed for body motion generalize to pure hand motion when trained on a large-scale, text-paired bimanual dataset. We adapt one representative model from each major generative family (diffusion, masked-token, and autoregressive) to our shared $\mathcal{R}_{\text{hand}}$ representation, train them on the RoMo Hands split, and score them under matched semantic and physical metrics.

\subsection{Evaluation Protocol}
\label{sec:eval_protocol}

We adopt the two-tier text-to-motion evaluation protocol of HumanML3D~\citep{humanml3d} and SnapMoGen~\citep{snapmogen}, which was later adapted by CLUTCH~\citep{clutch} for bimanual hands. This protocol comprises \emph{semantic} metrics scored by a learned text--motion evaluator, and \emph{physical} metrics computed directly on the generated MANO meshes. Full metric definitions and hyperparameters for the TMR~\citep{tmr} evaluator are provided in the supplementary material.

\paragraph{Semantic metrics.}
Following SnapMoGen~\citep{snapmogen}, we utilize a TMR evaluator trained on the RoMo Hands dataset. To ensure the evaluator scores articulation rather than spatial reconstruction, which would otherwise disproportionately reward VQ-style methods, we strip positions and velocities before normalization. We report: FID~\citep{fid}, measuring the distributional distance between generated and ground-truth (GT) motions; R-Precision (R@$k$)~\citep{humanml3d} for $k \in \{1,3,10\}$, measuring the retrieval accuracy of the correct caption from a pool of 32 captions (1 correct, 31 distractors drawn uniformly from the test set); MM-Dist, the average feature distance between a generated motion and its paired text; CLIP score, the cosine similarity between a generated motion and its paired text in the same TMR joint embedding space; Diversity, the variance across the generated set; and MultiModality, the variance of outputs generated from a single repeated prompt.

\paragraph{Physical metrics.}
To assess kinematic plausibility, we compute the following directly on the generated MANO meshes: Jerk ($\downarrow$), which quantifies the temporal smoothness of the motion; Inter-hand penetration ($\downarrow$), the percentage of frames where the left and right hand meshes intersect; Maximum self-hand penetration ($\downarrow$), the worst-case collision depth within a single hand (a depth-valued variant of the BimArt~\citep{bimart} self-penetration metric; see the supplementary material); and HF Power Ratio ($\downarrow$), the fraction of velocity energy residing in the high-frequency band. The HF Power Ratio is speed-normalized, isolating high-frequency jitter from the overall motion velocity to complement the absolute Jerk metric.

\paragraph{Test-set protocol.}
For each generator, we sample one motion per test prompt on the 17{,}348 RoMo Hands test clips of 40--196 frames, the training length range. Sequence lengths are cropped to the nearest multiple of 4 to accommodate VQ tokenizer temporal downsampling. We apply the same type-aware normalization used during training, extract the essential features, and embed them via the frozen TMR evaluator. During each replication, we draw the conditioning caption uniformly at random across the five granularity levels (L0--L4). Consequently, reported numbers are means marginalized over both replication noise and caption granularity. A per-level breakdown is available in the supplementary materials. Ground-truth motions are processed through the identical pipeline to establish the reference distribution. Because the split is segment-level, we also remove every test clip whose source video appears in train and recompute the benchmark. The four-model ranking is preserved, with R@1 shifts between $-0.026$ and $+0.009$; \tabref{tab:video_disjoint_audit} reports the full audit. We will additionally release a strict source-video-disjoint partition with the dataset.

\paragraph{Implementation details.}
\label{sec:impl}
We benchmark Kimodo M~\citep{kimodo} and MDM~\citep{mdm} (diffusion), MoMask~\citep{momask} (residual-VQ masked-token), and MotionMillion~\citep{motionmillion} (FSQ-tokenized autoregressive). All four were originally proposed for full-body motion; we retrain each on RoMo Hands under the shared $\mathcal{R}_{\text{hand}}$ representation, a frozen T5-v1.1-base text encoder~\citep{t5}, AdamW~\citep{adamw} optimization, and type-aware normalization, so any cross-architecture difference reflects representational fit rather than conditioning. Kimodo M is held at its published medium configuration (the only change is the input/output dimension from SMPL features to $\mathcal{R}_{\text{hand}}$); full per-model hyperparameters and schedules are in the supplementary materials.

\subsection{Results}
\label{sec:results_main}

We report semantic and physical results on the RoMo Hands test split in \tabref{tab:results_main}. Under the documented adaptations, the two metric families yield different rankings: MoMask and Kimodo M lead the semantic metrics, while no single model universally dominates the physical metrics. Because MDM is prefix-conditioned, MotionMillion is evaluated at 150M rather than its published 1B scale, and MoMask Stage~2 had not fully converged, per-model rankings should be read as indicative rather than definitive.

\begin{table}[t]
\centering
\caption{\textbf{No single generator dominates both metric families.} Semantic (left) and physical (right) results on the RoMo Hands test split. MoMask narrowly leads R-Precision retrieval, while Kimodo M leads the distribution-matching scores and most physical axes. ``$\rightarrow$'' marks metrics whose target is to \emph{match} GT (Diversity, Acceleration, Intra-hand Jerk). The italicized Ground Truth row is the reference distribution.}
\label{tab:results_main}
\setlength{\tabcolsep}{3pt}
\footnotesize
\resizebox{\linewidth}{!}{%
\begin{tabular}{@{} l ccc cc c | ccccc @{}}
\toprule
& \multicolumn{6}{c}{\textbf{Semantic}} & \multicolumn{5}{c}{\textbf{Physical}} \\
\cmidrule(lr){2-7} \cmidrule(lr){8-12}
Method
  & R@1\,$\uparrow$ & R@3\,$\uparrow$ & R@10\,$\uparrow$
  & MM-Dist\,$\downarrow$ & FID\,$\downarrow$ & Div.\,$\rightarrow$
  & \makecell{Accel $\rightarrow$\\(m/s$^2$)}
  & \makecell{Intra-Jerk $\rightarrow$\\(m/s$^3$)}
  & \makecell{Inter-Pen $\downarrow$\\(\%)}
  & \makecell{Self-Pen $\downarrow$\\(mm)}
  & \makecell{HF Pow. $\downarrow$\\(ratio)} \\
\midrule
\textit{Ground Truth} & \textit{63.77} & \textit{82.59} & \textit{96.13} & \textit{0.864} & \textit{0.000} & \textit{120.41} & \textit{8.38} & \textit{14.3} & \textit{0.64} & \textit{0.002} & \textit{0.0081} \\
\midrule
MDM (DiP)~\citep{mdm}               & 42.28 & 66.15 & 88.92 & 1.031 & 0.144 & 113.44 & 7.92 & 71.2  & 2.19 & 0.382 & 0.0132 \\
MoMask~\citep{momask}               & \textbf{54.00} & \textbf{78.62} & \textbf{93.88} & 0.945 & 0.129 & 115.27 & 4.99 & 265.2 & 1.58 & 0.017 & 0.0153 \\
Kimodo M~\citep{kimodo}             & 53.82 & 76.99 & 93.76 & \textbf{0.942} & \textbf{0.087} & \textbf{116.34} & 6.03 & 22.9  & \textbf{1.06} & 0.066 & \textbf{0.0077} \\
MotionMillion~\citep{motionmillion} & 31.26 & 51.55 & 82.37 & 1.105 & 0.314 & 109.14 & 5.77 & 76.1  & 1.82 & \textbf{0.000} & 0.0082 \\
\bottomrule
\end{tabular}}
\end{table}

\paragraph{Semantic results.}
MoMask (masked-token) and Kimodo M (diffusion) are statistically tied at the top of the semantic rankings: MoMask slightly edges out R-Precision retrieval, while Kimodo M leads on distribution-matching scores (FID, MM-Dist). MotionMillion and MDM (DiP) trail under their evaluated configurations; the latter uses a prefix-completion variant with a restricted 60-frame horizon.

\paragraph{Physical results.}
The physical landscape inverts the semantic ranking. Kimodo M is the cleanest overall with the lowest inter-hand penetration, the lowest HF power, and the best intra-hand jerk among baselines. MotionMillion incurs zero detected self-penetration and tracks GT HF power most closely, consistent with temporal regularization from its FSQ codebook. MoMask shows a counterintuitive pattern: its acceleration sits well \emph{below} GT yet its intra-hand jerk reaches $18\times$ GT, a slow-but-jagged signature consistent with abrupt residual-VQ token transitions that semantic metrics do not expose. MDM (DiP) has the highest contact violation rate under its evaluated 60-frame, 10-step configuration.

This split between metric families is a primary finding of the benchmark: \emph{no single evaluated configuration dominates both}. The preferred configuration depends on whether a downstream application prioritizes semantic text alignment or physical plausibility. This direction-level finding does not depend on treating the per-model ranking as definitive: pure-hand generation cannot be evaluated on semantic retrieval alone and requires complementary physical evaluation. The interactive statistics dashboard, side-by-side animated benchmark comparisons of every model on matched test prompts, and the 3D motion-grid visualizer are all available in the supplementary materials. We encourage the reader to inspect the comparisons in particular, since over-smoothing and high-frequency jitter are perceptually obvious in motion but only summarized in numbers here.

\begin{figure}[t]
  \centering
  \includegraphics[width=\linewidth]{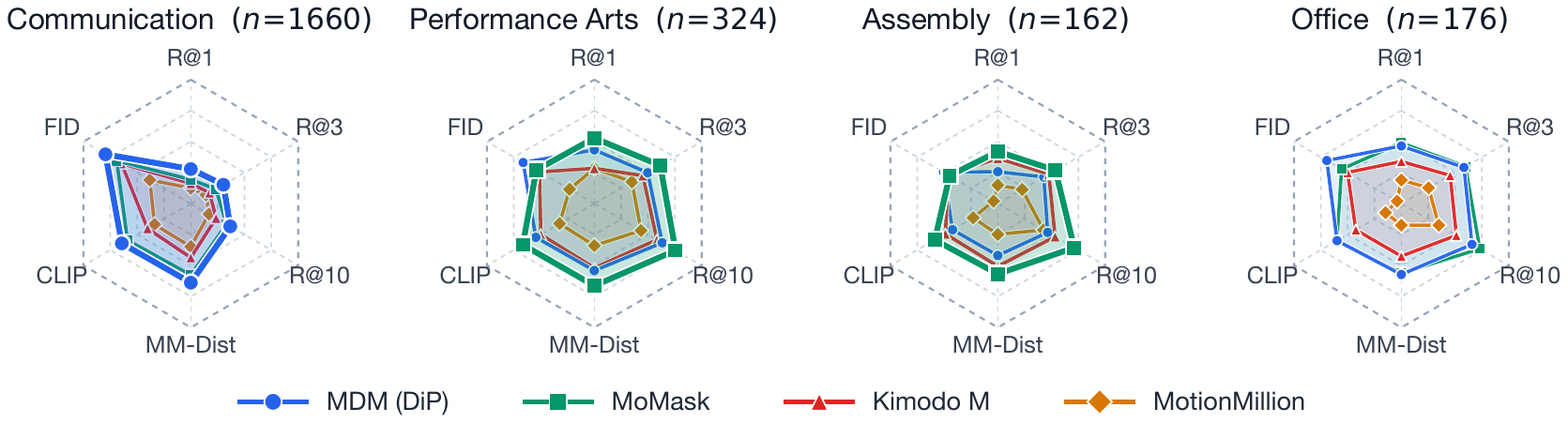}
  \caption{\textbf{Taxonomy uncovers per-category performance differences that are hidden in aggregate.} No single generator wins across the taxonomy: the leader rotates both \emph{across} categories and, in \emph{Office}, \emph{within} a single panel across the six semantic axes (R@1, R@3, R@10, MM-Dist, CLIP, FID), normalized to $[0,1]$. The supplementary covers all 40 categories of the pruned taxonomy.}
  \label{fig:taxonomy_radar}
\end{figure}

\paragraph{Taxonomy-stratified results.}
\Cref{fig:taxonomy_radar} shows the semantic metrics over the taxonomy categories and reveals blind spots that the aggregate \tabref{tab:results_main} hides. The leader rotates across categories: MDM (DiP) sweeps the \emph{Communication} category that has $n=1660$ samples; MoMask leads \emph{Performance Arts} ($n{=}324$) and \emph{Assembly} ($n{=}162$) on five of the six axes while MDM keeps the FID edge; and within \emph{Office} ($n{=}176$) the rotation persists across metric families, with MoMask winning the retrieval cluster (R@1, R@3, R@10, CLIP) and MDM winning the distribution-matching axes (MM-Dist, FID). MotionMillion does not lead any category by mean, consistent with its aggregate performance. The full breakdown over all 40 categories, with semantic and physical-metric stratification, is available in the supplementary materials.

\section{Conclusion}
\label{sec:conclusion}

\textbf{RoMo Hands} is an expansive in-the-wild dataset comprising $\sim$542K text-paired sequences across 884 hours. By pairing this scale with a five-level hierarchical captioning scheme and a coarse-to-fine taxonomy, we address the critical data bottleneck in dexterous motion synthesis. Our benchmark, underpinned by a novel 393-D bimanual representation ($\mathcal{R}_{\text{hand}}$), shows that body-motion generators exhibit distinct trade-offs under hand-specific adaptations. In particular, semantic and physical rankings diverge, and representation fit affects two different generator families. By releasing the dataset, taxonomy, and trained evaluator, we provide a robust foundation for the next generation of text-to-motion research.

\paragraph{Limitations.}
\label{sec:limitations}
Sourcing in-the-wild video introduces residual noise from occlusions and motion blur, which we mitigate via filtering and evaluate through a 250-clip human audit. The audit supports average caption fidelity and pose plausibility, but its five-point ratings are coarse and do not replace reconstruction error against ground-truth 3D hand pose. The dexterity filter introduces a selection bias toward well-observed, active motion rather than the raw web-video distribution, and the dataset inherits the demographic and geographic composition of its public source pool. Retrieval metrics use a single TMR evaluator trained on RoMo Hands, so they support relative comparisons under this shared evaluator rather than absolute comparison to body-motion benchmarks. Finally, the benchmark uses documented but non-identical adaptations---prefix conditioning for MDM, a 150M MotionMillion decoder, and a still-improving MoMask Stage~2---so architecture-specific rankings are indicative rather than definitive.

\paragraph{Future Work.}
Promising directions include scaling autoregressive decoders, such as  pushing the MotionMillion~\citep{motionmillion} decoder to 1B+ parameters, as well as exploring cross-dataset transfer to complementary benchmarks such as HandX~\citep{handx} and CLUTCH~\citep{clutch}. We also aim to integrate rich latent object context for hand-object interaction synthesis. Ultimately, this work paves the way for a unified multimodal generator capable of jointly synthesizing body, hands, and facial expressions under a shared tokenizer.

\paragraph{Broader Impact.}
RoMo Hands accelerates research in AR/VR avatars, telepresence, and robotics. To mitigate misuse, we release strictly de-identified motion features and text annotations. However, as generative hand models mature, they could theoretically be combined with synthesized visuals to create misleading gestural content. We urge the community to employ responsible access controls and usage policies when deploying downstream applications.

\clearpage

\medskip
\begingroup \small
\bibliographystyle{plainnat}
\bibliography{references}
\endgroup

\clearpage
\pagebreak
\newpage

\appendix

\begin{center}
{\Large\bfseries Supplementary Material for\par}
\vspace{0.3em}
{\large\bfseries \papertitle\par}
\end{center}
\vspace{1em}

\paragraph{Companion HTML files.}
A set of HTML files accompanies this PDF and is best viewed in a modern browser; each is self-contained and reproducible from the released dataset. We strongly recommend opening \texttt{index.html} first as it is a single-page portal to every artifact below, with short descriptions and direct entry points into the interactive dashboards, the 3D motion grid, and the side-by-side generator comparisons that complement the static figures and tables in this paper.
\begin{itemize}[leftmargin=*,topsep=2pt,itemsep=2pt]
  \item \texttt{index.html}: entry portal for the supplementary website; one card per artifact below with a short description and direct link.
  \item \texttt{RoMo\_Hands\_generative\_method\_comparison.html}: side-by-side animation playback of ground-truth, MotionMillion, Kimodo M, MDM, and MoMask outputs on matched train, validation, and test prompts.
  \item \texttt{Dataset\_statistics.html}: interactive statistical report covering durations, activity classes, per-finger and per-joint dynamic scores, inter-hand distance, smoothness, autocorrelation, and per-tier caption statistics, with cross-dataset comparison against GRAB and GigaHands. A curated subset of these views is reproduced as static figures in \appref{app:dataset_comparison_full}.
  \item \texttt{Hand\_motion\_explorer.html}: a searchable, filterable card grid over a 20-clip subset of RoMo Hands; each card opens a full MANO viewer with a timeline scrubber, the five-level captions, taxonomy labels, and per-clip quality scores.
  \item \texttt{Hand\_motion\_example\_grid.html}: a single-scene 3D viewer that plays the same 20 representative bimanual sequences in a synchronized grid; clicking a tile focuses it for scrubbing and inspection.
  \item \texttt{RoMo\_Hands\_generative\_method\_benchmark.html}: quantitative evaluation of every benchmark model, including taxonomy-stratified semantic retrieval (R@K, MM-Dist, CLIP), distributional metrics (FID, diversity), and physical plausibility.
\end{itemize}

\paragraph{Scope.}
This supplement expands the main paper along five axes: extended dataset comparisons including body-only datasets that did not fit in the main table (\appref{app:dataset_comparison_full}); the full specification of the bimanual hand motion representation $\mathcal{R}_{\text{hand}}$, including channel layout, type-aware normalization, and a comparison to alternative parameterizations (\appref{app:hand_repr}); precise mathematical definitions of every semantic and physical metric (\appref{app:metric_details}); the complete result tables with $95\%$ confidence intervals, additional baselines, and the per-caption-level Ground Truth breakdown (\appref{app:results_full}); and per-model implementation details, training schedules, and scaling decisions (\appref{app:impl}). The main paper stands alone; this supplement is optional further reading for reproducibility and depth.

\section{Full Dataset Comparison}
\label{app:dataset_comparison_full}

\begin{figure}[ht]
  \centering
  \begin{subfigure}[t]{0.58\textwidth}
    \centering
    \includegraphics[width=\linewidth]{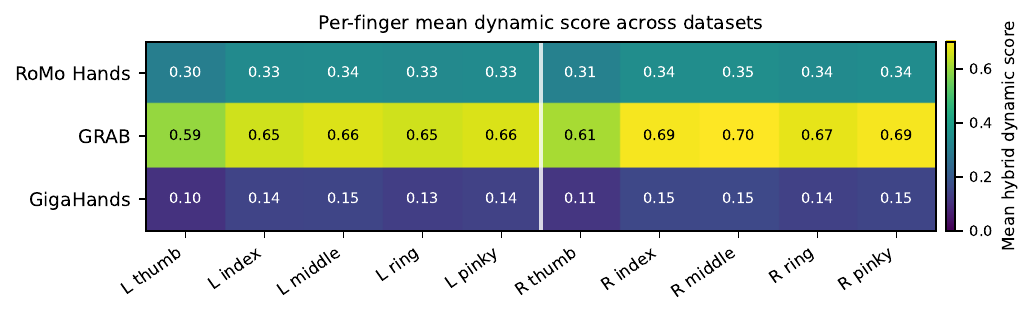}
    \caption{Per-finger mean hybrid dynamic score.}
    \label{fig:supp_finger_heatmap}
  \end{subfigure}
  \hfill
  \begin{subfigure}[t]{0.40\textwidth}
    \centering
    \includegraphics[width=\linewidth]{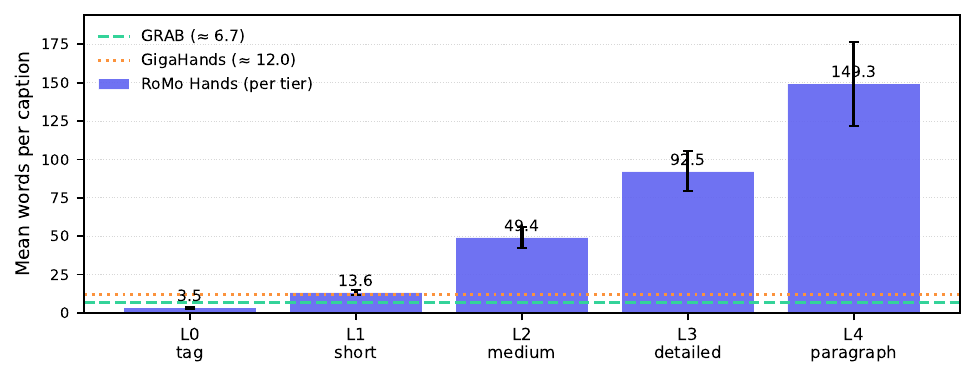}
    \caption{Caption-tier word counts.}
    \label{fig:supp_caption_tiers}
  \end{subfigure}
  \caption{\textbf{Finger coverage and caption-tier depth across datasets.} Left: GRAB saturates every finger near $0.65$, GigaHands' uncurated stream sits at $0.10$--$0.15$, and RoMo Hands occupies the middle ($0.30$--$0.35$) with thumbs slightly less dynamic than the long fingers (a consistent cross-dataset pattern). Right: RoMo Hands' five-tier caption hierarchy spans $3.5$ to $149.3$ words ($43\!\times$), while GigaHands ships a single caption per clip and the GRAB line shows our template sentences built from its object--action labels (GRAB itself is text-free).}
  \label{fig:supp_finger_caption}
\end{figure}

\begin{figure}[ht]
  \centering
  \includegraphics[width=\linewidth]{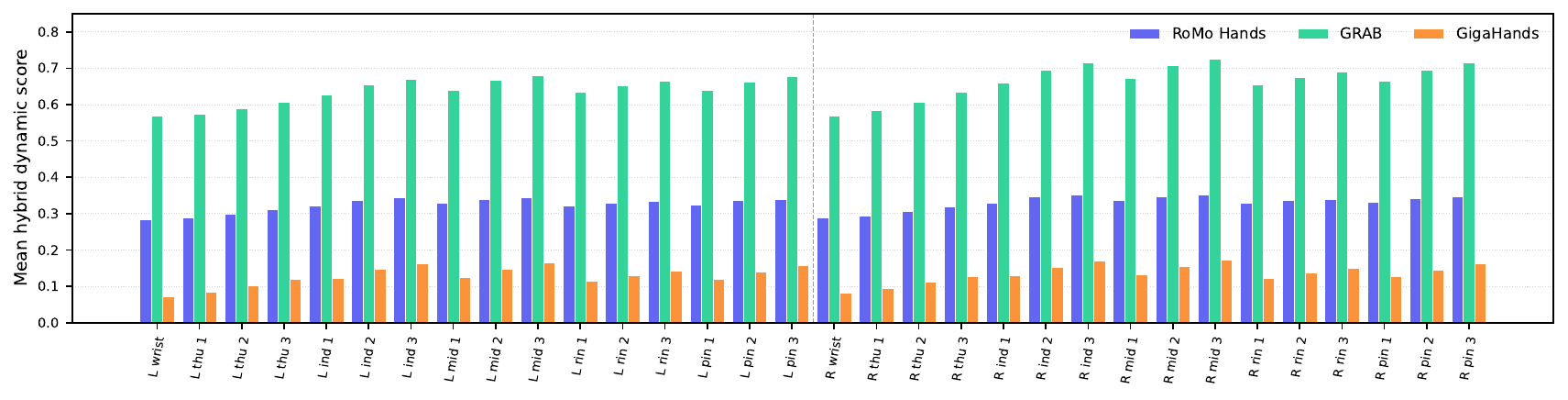}
  \caption{\textbf{Per-joint mean hybrid dynamic score} for all 32 MANO joints (16 left, 16 right; vertical guide marks the hand boundary). GRAB is uniformly saturated ($\approx\!0.65$); RoMo Hands shows the expected proximal-to-distal gradient (wrist $\approx\!0.28$, distal phalanges $\approx\!0.35$); GigaHands sits at $0.08$--$0.17$ across the entire skeleton. Joints are ordered wrist $\to$ thumb $\to$ index $\to$ middle $\to$ ring $\to$ pinky, with depth indices 1--3 from proximal to distal.}
  \label{fig:supp_joint_bar}
\end{figure}

This appendix expands the cross-dataset claims in \secref{sec:dataset_stats} along four axes: a wider table that situates RoMo Hands against the full body-motion regime in addition to the pure-hand and hand--object datasets shown in the main paper (\tabref{tab:dataset_comparison_full}); a like-for-like activity-coverage comparison against GRAB and GigaHands under a shared scoring pipeline (\tabref{tab:dataset_activity_comparison}); per-finger and per-joint dexterity coverage across the three datasets (\figref{fig:supp_finger_caption}, \figref{fig:supp_joint_bar}); and a caption-tier depth comparison highlighting RoMo Hands' five-level hierarchy (\figref{fig:supp_finger_caption}, right).

\paragraph{Body-scale context.}
The pure-hand datasets in \tabref{tab:comprehensive_dataset_comparison} top out at $\sim\!83$\,h (CLUTCH); the body-motion regime, by contrast, has scaled by roughly two orders of magnitude in the last three years, from HumanML3D's $28.6$\,h workhorse~\citep{humanml3d} to Motion-X ($144$\,h)~\citep{motionx}, MotionLib ($1456$\,h)~\citep{motionlib}, RoMo ($1238$\,h)~\citep{romo_body}, and MotionMillion ($2000$\,h)~\citep{motionmillion}. \tabref{tab:dataset_comparison_full} places RoMo Hands directly against this body-scale block: at $884$\,h, RoMo Hands is the first pure-hand dataset to reach the same band, an order of magnitude beyond the previous pure-hand state of the art and squarely inside the regime that recently unlocked the current generation of body-motion benchmarks. The body datasets themselves are not directly comparable on hand fidelity (they treat hands as rigid SMPL end-effectors), but their recent scale trajectory is the relevant yardstick for the data jump that RoMo Hands brings to the pure-hand regime.

\begin{table*}[!ht]
\centering
\caption{\textbf{RoMo Hands reaches body-motion scale in the pure-hand regime.} The body-only block (top) shows the recent two-orders-of-magnitude scale-up in body-motion datasets, from HumanML3D's $28.6$\,h to MotionMillion's $2000$\,h; pure-hand datasets (bottom) had remained below $100$\,h until now. RoMo Hands' $884$\,h closes that gap, sitting an order of magnitude beyond the previous pure-hand state of the art and inside the band that recently unlocked the current generation of body-motion benchmarks. Body datasets are not directly comparable on hand fidelity --- they treat hands as rigid SMPL end-effectors --- but their scale trajectory is the relevant yardstick. Hours and frame counts are the originally reported totals for each source dataset at its source frame rate; RoMo Hands is at 30\,fps.}
\label{tab:dataset_comparison_full}
\resizebox{\linewidth}{!}{
\begin{tabular}{@{} l l r r c r @{}}
\toprule
\textbf{Dataset} & \textbf{Scope} & \makecell{\textbf{Duration}\\\textbf{(h)}} & \makecell{\textbf{Frames}\\\textbf{(M)}} & \makecell{\textbf{Text}\\\textbf{Granularity}} & \makecell{\textbf{Text}\\\textbf{(K)}} \\
\midrule
HumanML3D \citep{humanml3d}    & Body (no hands)  & 28.6   & 2.1   & fine-grained & 45.0   \\
Motion-X \citep{motionx}       & Body (no hands)  & 144.2  & 15.6  & coarse       & 8.1    \\
InterAct \citep{interact}      & Body (no hands)  & 30.7   & 3.3   & coarse       & 48.6   \\
MotionLib \citep{motionlib}    & Body (no hands)  & 1456.4 & 157.3 & fine-grained & 2480.0 \\
MotionMillion \citep{motionmillion} & Body (no hands) & 2000.0 & 216.0 & fine-grained & 2000.0 \\
RoMo \citep{romo_body}         & Body (no hands)  & 1237.8 & 133.7 & fine-grained & 4100.0 \\
\midrule
GRAB \citep{grab}              & Hands + Object   & 3.8    & 1.6   & ---          & ---    \\
OakInk2 \citep{oakink2}        & Hands + Object   & 37.1   & 4.01  & action       & ---    \\
ARCTIC \citep{arctic}          & Hands + Object   & 2.02   & 0.22  & action       & ---    \\
H2O \citep{h2o}                & Hands + Object   & 1.06   & 0.11  & action       & ---    \\
HOT3D \citep{hot3d}            & Hands + Object   & 3.90   & 0.42  & ---          & ---    \\
GigaHands \citep{gigahands}    & Hands + Object   & 34.0   & 3.7   & coarse       & 84.0   \\
HoloAssist \citep{holoassist}  & Hands + Object   & 161.2  & 17.4  & coarse       & 1.8    \\
\midrule
HandDiffuse \citep{handdiffuse}      & Pure Hands & 2.0    & 0.25  & ---          & ---    \\
BOTH2Hands \citep{both2hands}        & Pure Hands & 8.31   & 1.8   & coarse       & 23.5   \\
InterHand2.6M \citep{interhand}      & Pure Hands & 24.0   & 2.6   & ---          & ---    \\
HandX \citep{handx}                  & Pure Hands & 54.2   & 5.9   & fine-grained & 485.7  \\
CLUTCH (3D-HIW) \citep{clutch}       & Pure Hands & 83.3   & 12.0  & fine-grained & 32.0   \\
\midrule
\rowcolor{blue!5}
\textbf{RoMo Hands (Ours)} & \textbf{Pure Hands} & \textbf{884}   & \textbf{95.5}  & fine-grained & \textbf{2710}  \\
\bottomrule
\end{tabular}
}
\end{table*}

\paragraph{Dynamic score.}
We score per-clip motion activity with a \emph{hybrid dynamic score} $\delta_j$ that combines a temporal term and a spatial term in equal weight,
\[
  \delta_j \;=\; \tfrac{1}{2}\,\overline{\|v_{t,j}\|_2}^{\,t} \;+\; \tfrac{1}{2}\,\bigl\| \max_t p_{t,j} \,-\, \min_t p_{t,j} \bigr\|_2,
\]
where the first term is the clip-mean speed of joint $j$ in metres per second and the second is the $\ell_2$ norm of the per-axis bounding-box extents of joint $j$ over the clip in metres.
Both terms are kept in their native physical units (no sigmoid, no per-dataset percentile cap, and no cross-term rescaling), so $\delta_j$ is a mixed-unit summary in which the spatial term tends to dominate on typical bimanual hand clips.
Joint positions are taken in the bimanual-canonical frame of $\mathcal{R}_{\text{hand}}$ (root-relative $xz$ in the inter-wrist heading frame, world-absolute $y$; \secref{app:hand_repr_breakdown}), so $\delta$ is invariant to where the subject is in the scene and which way they are facing, and measures motion relative to the bimanual root.
Joints aggregate arithmetically: a per-finger score averages the three phalanges of that finger (no fingertips, $5$ fingers $\times\,2$ hands $=10$ groups), a per-hand score averages its $15$ finger joints, and the per-clip ``overall'' score averages the $30$ finger joints across both hands with the two wrists excluded; wrist scores are tracked separately over $\{0, 16\}$.
\figref{fig:supp_finger_caption} and \figref{fig:supp_joint_bar} report the per-finger and per-joint hybrid score respectively (the bar chart includes wrists; the per-finger heatmap and the headline ``Mean Dynamic'' do not), and \tabref{tab:dataset_activity_comparison}'s Mean Dynamic column is the across-clips mean of each clip's overall hybrid score.

\paragraph{Activity ratio and buckets.}
A complementary, simpler statistic is the per-clip \emph{activity ratio}: the fraction of frames whose mean joint velocity exceeds $0.01$\,m/s. This ratio assigns each clip to one of three buckets, Static ($<\!10\%$), Low ($10$--$50\%$), or Active ($>\!50\%$), which are the columns reported in \tabref{tab:dataset_activity_comparison}; the Mean Dynamic column on the same table reports the mean of $\delta$ across all clips in each dataset.

\paragraph{Activity coverage under shared scoring.}
\tabref{tab:dataset_activity_comparison} reports the cross-dataset comparison summarized in \secref{sec:dataset_stats}: each clip in GRAB, GigaHands, and RoMo Hands is scored by the same dynamic-score function defined above, then bucketed under shared thresholds. We do \emph{not} re-segment GRAB or GigaHands; their clips are taken as released by the original authors and resampled to our 30\,fps reference rate, with clips that fail to load discarded, and only the scoring tool is shared, so per-clip scores are directly comparable across the three datasets without altering the reference splits. (The cross-dataset totals in \tabref{tab:comprehensive_dataset_comparison} and \tabref{tab:dataset_comparison_full} preserve each source paper's originally reported sequence count, hours, and frames at the source frame rate; the activity-coverage analysis below is reported under our 30\,fps loader and therefore lists 2.86\,h / 16.40\,h for GRAB / GigaHands respectively.) Duration histograms are normalized to density so shapes are comparable across very different dataset sizes. Three patterns stand out: \textbf{scale}: $541{,}929$ clips (884\,h) vs. GRAB's $1{,}067$ (2.86\,h) and GigaHands' $12{,}704$ (16.4\,h), 2--3 orders of magnitude more clips at comparable per-clip lengths; \textbf{curation effect}: GigaHands' uncurated egocentric stream is $85.8\%$ Static+Low ($20.4\%$ Static), whereas RoMo Hands collapses Static to $0.2\%$ and recovers a $53.7\%$ Active majority despite being two orders of magnitude larger; \textbf{regime placement}: the RoMo Hands mean dynamic score ($0.330$) sits between GRAB's tightly choreographed studio capture ($0.658$) and GigaHands' uncurated stream ($0.136$), RoMo Hands trades GRAB's onset-saturated grasp-cycle regime for broader coverage while still removing the static mean-pose mass that dominates raw web video, and we unpack the per-finger and per-joint structure of these bands below.

\begin{table*}[!ht]
\centering
\caption{\textbf{Cross-dataset activity coverage under a shared scoring tool.} Each clip in GRAB, GigaHands, and RoMo Hands is scored by the same per-frame dynamic-score function used during RoMo Hands curation. Activity buckets (Static / Low / Active) are assigned per clip from shared thresholds; Mean Dynamic is the average per-clip score in $[0,1]$.}
\label{tab:dataset_activity_comparison}
\resizebox{\linewidth}{!}{
\begin{tabular}{lrrrrcccc}
\toprule
\textbf{Dataset} & \textbf{Sequences} & \textbf{Total (h)} & \textbf{Mean (s)} & \textbf{Median (s)} & \textbf{Active (\%)} & \textbf{Low (\%)} & \textbf{Static (\%)} & \textbf{Mean Dynamic} \\
\midrule
GRAB \cite{grab}              & 1{,}067   & 2.86   & 9.66 & 8.10 & 65.6 & 34.4 & 0.0  & 0.6579 \\
GigaHands \cite{gigahands}    & 12{,}704  & 16.40  & 4.65 & 3.13 & 14.2 & 65.4 & 20.4 & 0.1357 \\
\midrule
\rowcolor{blue!5}
\textbf{RoMo Hands (Ours)}    & \textbf{541{,}929} & \textbf{884.17} & 5.87 & 4.00 & \textbf{53.7} & \textbf{46.1} & \textbf{0.2}  & \textbf{0.3304} \\
\bottomrule
\end{tabular}
}
\end{table*}

\paragraph{Per-finger and per-joint coverage.}
\figref{fig:supp_finger_caption} reports the mean hybrid dynamic score per finger across the three datasets and \figref{fig:supp_joint_bar} reports the same quantity at the per-joint level ($32$ MANO joints). The three datasets occupy cleanly separated bands with GRAB at $0.57$--$0.72$, RoMo Hands at $0.28$--$0.35$, GigaHands at $0.07$--$0.17$, which we read as diagnostic of \emph{collection setup} rather than as an intrinsic ordering of motion quality: $\delta$ is a sum of physical units rather than a normalized score, so a higher value here is not automatically ``better.''
GRAB sits at the top because every clip is a short studio-recorded grasp + manipulate + release cycle: the rest-to-grasp transition alone sweeps the fingers across most of their workspace, which inflates the spatial bounding-box term every clip even before the manipulation phase, and the absence of idle frames in between keeps the velocity term elevated as well.
GigaHands sits at the bottom because its uncurated egocentric stream contains substantial idle / hands-resting frames: with hands at rest, both the per-joint bounding box and the average speed collapse, and a per-clip average over a stream where most frames look like that necessarily lands in the low-dynamic regime.
RoMo Hands lands in the middle by design: dexterity filtering at curation removes the static frames that depress GigaHands, while the absence of a forced rest-to-grasp framing avoids the onset-driven inflation that lifts GRAB.
Within the RoMo Hands band the figures recover two anatomical patterns one would expect: a proximal-to-distal gradient driven by the bimanual-canonical frame (wrists $0.28$--$0.29$, where motion in the canonical frame collapses to inter-wrist-distance variation plus world-$y$ displacement, vs.\ distal phalanges $0.32$--$0.35$ where the full articulation workspace is exposed) and slightly less active thumbs than long fingers (thumbs $\approx 0.30$, index--pinky $0.33$--$0.35$), confirming that the score still discriminates major from minor articulators on clip distributions where neither collection-setup artefact dominates. These figures are a static slice of the interactive views in \texttt{Dataset\_statistics.html}.

\paragraph{Caption-tier word counts.}
\figref{fig:supp_finger_caption} (right) plots the mean number of words per caption across the five RoMo Hands caption tiers ($L_0$ tag $\to$ $L_4$ paragraph) against the comparators' single-caption baselines. RoMo Hands grows from $3.5$ words at $L_0$ to $149.3$ at $L_4$  (nearly a $43\!\times$ span across tiers) whereas GigaHands ships a single, fixed-granularity caption per clip ($\sim\!12.0$ words). GRAB releases no captions; its reference line ($\sim\!6.7$ words) shows template sentences we generate from GRAB's object and action labels (e.g., ``a person drinks from a mug''), included for reference only. This explicit hierarchy is what enables the per-tier retrieval breakdown of \tabref{tab:gt_caption_breakdown_app} and the uniform-over-levels training mixture used by the TMR evaluator.

\section{Bimanual Hand Motion Representation: Full Specification}
\label{app:hand_repr}

This appendix expands the compact main-text description in \secref{sec:hand_repr}.
We start from MANO and the body-motion-generation template, formalize the channel equivalence between $\mathcal{R}_{272}$~\citep{motionstreamer,motionmillion,lixingxiao272repr} and our $\mathcal{R}_{\text{hand}}$, list the per-dimension layout, derive the type-aware normalization scheme, and compare to alternative bimanual parameterizations.

\subsection{From MANO to a body-style representation}
\label{app:hand_repr_mano}

MANO~\citep{mano} parameterizes each hand by a global axis-angle rotation $\mathbf{r}^g\!\in\!\mathbb{R}^3$, a translation $\mathbf{t}^g\!\in\!\mathbb{R}^3$, $15$ local joint rotations ($\mathbb{R}^{45}$), and $10$ PCA shape coefficients $\boldsymbol{\beta}\!\in\!\mathbb{R}^{10}$ (61 per hand, 122 bimanually). Existing hand generators~\citep{clutch} drop $\boldsymbol{\beta}$ and convert rotations to a 6D continuous form~\citep{zhou2019continuity}, yielding the 99-D CLUTCH parameterization.

Body-motion generators consume a richer template: the HumanML3D 263-D~\citep{humanml3d} and 272-D~\citep{motionstreamer,motionmillion,lixingxiao272repr} representations pack rotations alongside redundant root-relative joint positions, finite-difference velocities, and a 6D heading frame, and this redundancy is what makes diffusion, masked-token, and autoregressive backbones converge well. Our $393$-D representation transports this template to MANO's 32-joint bimanual skeleton (16 joints per hand; a wrist root and five three-phalanx finger chains, indices $0$--$15$ left, $16$--$31$ right), uses 6D rotations throughout, and drops $\boldsymbol{\beta}$ since motion is generated at fixed mean shape. Two decisions distinguish it from a naive port:
\begin{enumerate}[label=(\roman*),leftmargin=*,topsep=2pt,itemsep=2pt]
\item \textbf{Heading is derived from the inter-wrist axis.} Body
representations canonicalize a clip by aligning a pelvis-anchored heading;
hands have no natural pelvis. We extract a yaw angle $\psi_t$ from the
wrist-to-wrist direction and apply $\mathbf{R}_y(\psi_t)$, a rotation about
the world-up axis only. Restricting heading canonicalization to yaw preserves
the absolute $y$-component of every position, exposing world height as a
directly observable feature and avoiding the gravity-coupled rotations that
would otherwise corrupt joint-position channels for two-handed gestures.
\item \textbf{The bimanual root is the wrist midpoint, not a skeleton joint.}
We define $\mathbf{r}_t = \tfrac{1}{2}(\mathbf{p}^L_t + \mathbf{p}^R_t)$ as
the root used by the heading frame and the velocity channel. Because
$\mathbf{r}_t$ is not one of the $32$ skeleton joints, its height is stored as
an explicit scalar; without this, root height could not be recovered without
integration.
\end{enumerate}

\subsection{Structural equivalence to the body 272-D representation}
\label{app:hand_repr_272}

The body $272$-D representation introduced by~\citet{motionstreamer} (adopted by~\citet{motionmillion}, with public reference implementation by~\citet{lixingxiao272repr}), hereafter $\mathcal{R}_{272}$, lays out five per-frame blocks (root XZ linear velocity, root angular velocity as a $6$D continuous rotation, and per-joint positions, velocities, and $6$D rotations over the $22$-joint SMPL skeleton), whose exact dimension ranges are listed in the body column of \tabref{tab:body_hand_template}.
Its only structural specialization is the absence of an explicit root-height channel: the pelvis is joint $0$ of the SMPL skeleton, so its world-$y$ component is carried inside the joint-position block via XZ-centering.
Stripping this one implicit-vs-explicit difference, $\mathcal{R}_{272}$ and our $\mathcal{R}_{\text{hand}}$ instantiate the same generic per-frame template $\mathbf{m}_t$ at $J\!=\!22$ and $J\!=\!32$ respectively, with the hand-specific vector $\mathbf{h}_t$ of \eqref{eq:hand_repr_template} recovered as the $J\!=\!32$ instance of $\mathbf{m}_t$:
\begin{equation}
  \mathbf{m}_t \;=\; \bigl[\,
  \underbrace{\mathbf{f}_t}_{\text{root frame}}\,,\;
  \underbrace{\{\tilde{\mathbf{p}}^j_t\}_{j=0}^{J-1}}_{3J,\;\text{joint positions}}\,,\;
  \underbrace{\{\dot{\tilde{\mathbf{p}}}^j_t\}_{j=0}^{J-1}}_{3J,\;\text{joint velocities}}\,,\;
  \underbrace{\{\phi(\mathbf{R}^j_t)\}_{j=0}^{J-1}}_{6J,\;\text{6D rotations}}
  \,\bigr],
  \label{eq:body_hand_template}
\end{equation}
with $\dim \mathbf{f}_t = 8$ when the root joint coincides with a skeleton joint (body) and $\dim \mathbf{f}_t = 9$ when the root is off-skeleton (our hand case, where the bimanual root is the wrist midpoint and an explicit height channel is required).
This gives $\dim \mathbf{m}_t = 12J + 8$ for the body case (with $J\!=\!22$ this recovers $\mathcal{R}_{272}$ exactly: $12\!\cdot\!22 + 8 = 272$) and $\dim \mathbf{m}_t = 12J + 9$ for the bimanual hand case (with $J\!=\!32$: $12\!\cdot\!32 + 9 = 393$).
\tabref{tab:body_hand_template} pins this equivalence channel-by-channel against the verified $\mathcal{R}_{272}$ index ranges.

\begin{table}[ht]
\caption{\textbf{$\mathcal{R}_{\text{hand}}$ is structurally the body $\mathcal{R}_{272}$ template re-indexed by joint count, with one extra channel forced by the off-skeleton bimanual root.}
Per-channel equivalence between the body $272$-D representation $\mathcal{R}_{272}$~\citep{motionstreamer,motionmillion,lixingxiao272repr} and the proposed hand $393$-D representation. Both instantiate the template of \cref{eq:body_hand_template}; the bimanual root being off-skeleton forces an explicit root-height channel on the hand side.}
\label{tab:body_hand_template}
\centering
\small
\begin{tabular}{@{}llcccc@{}}
\toprule
\textbf{Channel group} & \textbf{Channel content} &
\multicolumn{2}{c}{\textbf{Body} ($\mathcal{R}_{272}$, $J\!=\!22$)} &
\multicolumn{2}{c}{\textbf{Hand} ($\mathcal{R}_{\text{hand}}$, $J\!=\!32$)} \\
\cmidrule(lr){3-4}\cmidrule(lr){5-6}
                 &                                              & dim & range          & dim & range          \\
\midrule
Root frame       & root XZ linear velocity                      & 2   & $[0{:}2)$      & 2   & $[0{:}2)$      \\
                 & root height (absolute world $y$)             & --  & implicit       & 1   & $[2{:}3)$      \\
                 & root angular velocity (6D rotation)          & 6   & $[2{:}8)$      & 6   & $[3{:}9)$      \\
\addlinespace[2pt]
Joint positions  & root-relative, $J\!\times\!3$                & 66  & $[8{:}74)$     & 96  & $[9{:}105)$    \\
Joint velocities & finite-difference, $J\!\times\!3$            & 66  & $[74{:}140)$   & 96  & $[105{:}201)$  \\
Joint rotations  & 6D continuous~\citep{zhou2019continuity}, $J\!\times\!6$
                                                                & 132 & $[140{:}272)$  & 192 & $[201{:}393)$  \\
\midrule
\textbf{Total} & & \textbf{272}  & ($12J\!+\!8$) & \textbf{393} & ($12J\!+\!9$) \\
\bottomrule
\end{tabular}
\end{table}

\subsection{Detailed per-dimension breakdown}
\label{app:hand_repr_breakdown}

We now specify the exact layout of $\mathbf{h}_t \!\in\!\mathbb{R}^{393}$.
At frame $t$, let $\mathbf{p}^L_t, \mathbf{p}^R_t \!\in\!\mathbb{R}^3$ denote the wrist positions and $\mathbf{r}_t = \tfrac{1}{2}(\mathbf{p}^L_t + \mathbf{p}^R_t)$ the bimanual root.
The heading-frame yaw is
\begin{equation}
  \psi_t \;=\; \operatorname{atan2}\!\bigl(-(\mathbf{p}^R_t - \mathbf{p}^L_t)_z,\;
  (\mathbf{p}^R_t - \mathbf{p}^L_t)_x\bigr),
  \qquad
  \mathbf{R}^\psi_t \;=\; \mathbf{R}_y(\psi_t)\;\in\;SO(3).
\end{equation}
\paragraph{Heading degeneracy.}
The yaw $\psi_t$ is undefined whenever the two wrists project to the same point on the $XZ$ plane, i.e.\ whenever $(\mathbf{p}^R_t - \mathbf{p}^L_t)_x = (\mathbf{p}^R_t - \mathbf{p}^L_t)_z = 0$; this covers prayer, overhead-clasp, and the limit case of coincident wrists.
On each maximal degenerate run we replace $\psi_t$ by linear interpolation between the bracketing non-degenerate frames; if a run touches a clip boundary we hold the nearest non-degenerate value across it.
Because linear interpolation in $\psi$ keeps every replaced frame on the yaw subgroup $\{\mathbf{R}_y(\psi)\}_{\psi \in \mathbb{R}}$, the round-trip property of the type-aware normalization (\secref{app:hand_repr_normalization}) is preserved.

We store the heading rotation as a $6$D continuous descriptor $\boldsymbol{\omega}_t = \phi(\mathbf{R}^\psi_t)$, where the operator $\phi$ returns the first two columns of the rotation matrix flattened to $\mathbb{R}^6$~\citep{zhou2019continuity}.
By construction this descriptor encodes the absolute heading at the start of each clip and incremental yaw at all subsequent frames.
Joint quantities are expressed in the heading-canonical reference: $\tilde{\mathbf{p}}^j_t$ are joint positions made root-relative on the $XZ$ plane (their $y$-component is absolute world height) and yaw-aligned by $\mathbf{R}^\psi_t$; $\dot{\tilde{\mathbf{p}}}^j_t$ are their first-order finite differences; $\mathbf{R}^j_t \!\in\!
SO(3)$ are joint rotations, where the two wrist rotations ($j\!\in\!\{0,16\}$) are expressed in the heading-canonical frame and all finger rotations remain parent-relative.
\tabref{tab:hand_393_layout} reports the resulting layout.

\begin{table}[ht]
\caption{Per-dimension layout of the proposed hand $393$-D representation. The
type column drives the type-aware normalization of
\secref{app:hand_repr_normalization}: \textsc{gaussian} dims are $z$-scored
and clipped to $[-5,+5]\sigma$; \textsc{cyclic} dims pass through untouched
to preserve the Stiefel-manifold structure; \textsc{constant} dims are
zeroed on forward and restored from $\mu$ on inverse.}
\label{tab:hand_393_layout}
\centering
\resizebox{\textwidth}{!}{%
\begin{tabular}{@{}clrl@{}}
\toprule
\textbf{Dim range} & \textbf{Channel} & \textbf{Width} & \textbf{Type} \\
\midrule
$0$--$1$     & Root XZ velocity, heading-aligned ($\dot{\mathbf{r}}^{xz}_t$)        & 2   & \textsc{gaussian} \\
$2$         & Root height ($r_{t,y}$, absolute world $y$)                          & 1   & \textsc{gaussian}      \\
$3$--$8$     & Heading 6D rotation ($\boldsymbol{\omega}_t = \phi(\mathbf{R}_y(\psi_t))$) & 6   & \textsc{cyclic} ($\times 2$), \textsc{constant} ($\times 4$) \\
$9$--$104$   & Joint positions, root-relative \& heading-aligned ($\{\tilde{\mathbf{p}}^j_t\}$, $32\!\times\!3$) & 96  & \textsc{gaussian}      \\
$105$--$200$ & Joint velocities, finite difference ($\{\dot{\tilde{\mathbf{p}}}^j_t\}$, $32\!\times\!3$)         & 96  & \textsc{gaussian} \\
$201$--$392$ & Joint 6D rotations on $\operatorname{St}(3,2)$ ($\{\phi(\mathbf{R}^j_t)\}$, $32\!\times\!6$)      & 192 & \textsc{cyclic}     \\
\midrule
\multicolumn{2}{l}{\textbf{Total}} & \textbf{393} &  \\
\multicolumn{2}{l}{\quad of which \textsc{gaussian}}     & 195 & root XZ vel (2) + height (1) + positions (96) + joint vel (96) \\
\multicolumn{2}{l}{\quad of which \textsc{cyclic}}       & 194 & heading (2 of 6) + joint rotations (192) \\
\multicolumn{2}{l}{\quad of which \textsc{constant}}     & 4   & 4 of 6 heading dims (zero by construction) \\
\bottomrule
\end{tabular}}
\end{table}

\subsection{Choice of normalization scheme}
\label{app:hand_repr_normalization}

Two normalization schemes are natural candidates for $\mathbf{h}_t \!\in\!\mathbb{R}^{393}$: the \textbf{vanilla per-channel $z$-score} that is the de-facto standard on full-body data~\citep{humanml3d, motionmillion}, and the \textbf{type-aware} scheme we adopt. The vanilla recipe is empirically misapplied to $\mathbf{h}_t$ for three independent reasons rooted in geometry, not data quality; the type-aware scheme is a strict generalization that coincides with vanilla on a benign representation but diverges on $\mathbf{h}_t$.

\paragraph{Option A: vanilla per-channel $z$-score.}
Fit $(\mu_i, \sigma_i)$ on the train split and apply the same affine map per dimension,
\begin{equation}
  \tilde{h}_{t,i} \;=\; (h_{t,i} - \mu_i)/\sigma_i,
  \qquad \sigma_i \!\leftarrow\! \max(\sigma_i, 1),
  \label{eq:zscore}
\end{equation}
with the standard $\max(\sigma,1)$ guard against degenerate channels. This works on $263$-D and $272$-D body representations because every dimension is approximately Gaussian, well-conditioned, and geometrically unconstrained. None of those preconditions holds on $\mathbf{h}_t$:

\begin{enumerate}[label=(\roman*),leftmargin=*,topsep=2pt,itemsep=2pt]
\item \emph{Manifold-constrained channels.}
Each joint's $6$D rotation is not six free numbers but two $3$-vectors that must remain unit-norm and mutually orthogonal; rescaling the six coordinates independently breaks both constraints, so the encoder no longer sees rotations as rotations.
Concretely, the $6$D rotation features ($\boldsymbol{\omega}_t$ and
$\{\phi(\mathbf{R}^j_t)\}$, $192\!+\!2 = 194$ active dimensions) live on the
Stiefel manifold $\operatorname{St}(3,2)$ (the first two columns of a
rotation matrix), carrying $3\!\cdot\!32 + 1 = 97$ true degrees of freedom in
their $194$ ambient dimensions. The redundancy is the price of continuity:
\citet{zhou2019continuity} prove that no continuous mapping
$\mathbb{R}^n \!\to\! SO(d)$ exists below a critical $n$, so the constraint
$\|\mathbf{a}_1\|=\|\mathbf{a}_2\|=1$, $\mathbf{a}_1\!\cdot\!\mathbf{a}_2=0$
must be borne in the representation itself.
After $z$-scoring this constraint no longer holds in the input the encoder sees, and the manifold structure has to be re-discovered from training samples rather than being baked into the representation.

\item \emph{Bimodal channels by construction.}
Several channels mix two regimes with very different scales (a one-shot draw at the start of each clip versus a near-identity increment everywhere else, or the dual of that for velocities, where rare bursts coexist with long near-stationary runs), so a single $(\mu, \sigma)$ fits neither, and after normalization the rare regime surfaces as a $30$--$50\sigma$ ``outlier'' that is in fact an entirely valid frame.
The clearest case is the heading rotation $\boldsymbol{\omega}_t$, which stores the \emph{absolute}
heading at $t\!=\!0$ (uniformly distributed on $S^1$ across clips) and the \emph{incremental} change at $t\!\geq\!1$ (sharply peaked near identity, $\cos\Delta\psi\!\approx\!1,\;\sin\Delta\psi\!\approx\!0$). Per-channel statistics average over both regimes: in our training set the cached $\sigma$ for $\cos\psi$ is $0.055$, while the
$\mathrm{MAD}\!\cdot\!1.4826$ measured on frames with $t\!\geq\!1$ alone is
$4\!\times\!10^{-5}$ (approximately $1300\times$ smaller), so frame-zero values normalize as $30$--$50\sigma$ ``outliers'' even though they are perfectly valid orientations.
The dual pathology applies to the very-low-variance translation channels (root XZ velocity, joint velocities; $\sim\!98$ dims) on hand data, where most frames are near-stationary and $\sigma$ is dominated by rare bursts; the result is occasional normalized values exceeding $\pm 30$ on real validation data.

\item \emph{Constant-by-construction channels.}
Four of the six heading dimensions carry no information at all (they are pinned to $0$ or $1$ for every frame of every clip) yet the standard $\sigma\!\leftarrow\!\max(\sigma,1)$ guard rescales them as if they were random and the model has to learn that they are meaningless.
Specifically, these are the third row of $\mathbf{R}_y(\psi_t)$ and the second
column's $y$-component, which the yaw-only restriction pins by construction; their empirical $\sigma$ collapses to $0$, the $\max(\sigma,1)$ guard kicks in, and the resulting channels become structured noise that the encoder has to explicitly learn to ignore.
\end{enumerate}
None of these failure modes is data-driven: each follows from how $\mathbf{h}_t$ is constructed, and each would persist under perfect mocap.

\paragraph{Option B: type-aware normalization (used).}
We address all three pathologies with a per-dimension type mask $\tau\!\in\!\{\textsc{gaussian}, \textsc{cyclic}, \textsc{constant}\}^{393}$ (right-most column of \tabref{tab:hand_393_layout}) that selects one of three transforms:
\begin{equation}
  \tilde{h}_{t,i} \;=\;
  \begin{cases}
    \operatorname{clip}\!\bigl((h_{t,i} - \mu_i)/\sigma_i,\,\pm\sigma_{\text{clip}}\bigr), & \tau_i = \textsc{gaussian} \\
    h_{t,i},                                                              & \tau_i = \textsc{cyclic} \\
    0,                                                                    & \tau_i = \textsc{constant}
  \end{cases}
  \label{eq:typeaware}
\end{equation}
with $\sigma_{\text{clip}}\!=\!5$. The \textsc{cyclic} block covers all $194$ active rotation dims and is passed through, preserving the Stiefel-manifold structure (neutralizing~(i)). The \textsc{constant} block covers the four degenerate heading dims; they are zeroed on forward and restored from $\mu_i$ on inverse, so the round-trip is lossless (neutralizing~(iii)). The $195$ \textsc{gaussian} channels are $z$-scored and clipped at $\pm\sigma_{\text{clip}}$; the clip leaves the well-behaved position/height channels untouched (their values lie within $\pm 3\sigma$) and acts only on the $\sim\!98$ small-$\sigma$ velocity channels, absorbing the heavy-tail bursts of~(ii). $98.7\%$ of velocity-channel mass lies within $\pm 3\sigma$ on the training set, so the clip is conservative.

\paragraph{Relationship between the two options.}
Option~B reduces to Option~A as every dim is tagged \textsc{gaussian} and $\sigma_{\text{clip}}\!\to\!\infty$: \eqref{eq:typeaware} collapses to \eqref{eq:zscore}. The two schemes are operationally identical on benign representations like the body 272-D template, but diverge on $\mathbf{h}_t$ along the three axes above. The divergence matters empirically: the TMR retrieval evaluator trained under Option~A produces validation features with occasional $\pm 30\sigma$ excursions and degraded R@1, whereas the same architecture trained under Option~B converges to the noise floor reported in \secref{app:impl_tmr}. The transform is invertible per type and lossless on the \textsc{cyclic} block, preserving the full bimanual geometry end-to-end through training and evaluation. All hand experiments in this paper including the TMR evaluator are trained under Option~B with $\sigma_{\text{clip}}\!=\!5$.

\subsection{Comparison to prior bimanual parameterizations}
\label{app:hand_repr_compare}

\tabref{tab:hand_repr_compare} compares $\mathcal{R}_{\text{hand}}$ to the principal bimanual alternatives.
Our representation is roughly $2\!\times$ the dimensionality of CLUTCH's parameterization and $4\!\times$ that of raw MANO, but this redundancy follows the same principle that makes the body $263$-D and $272$-D representations effective for full-body generation: making geometric structure explicitly accessible through complementary views of the same underlying pose.

\begin{table}[ht]
\caption{\textbf{$\mathcal{R}_{\text{hand}}$ extends raw MANO and CLUTCH's parameterization with body-style position, velocity, and heading channels.} This roughly doubles the dimensionality but matches the inductive bias of HumanML3D-272. Rows enumerate the design axes that distinguish the three options; the redundancy of $\mathcal{R}_{\text{hand}}$ is the same redundancy that makes body templates effective for full-body generation.}
\label{tab:hand_repr_compare}
\centering
\small
\begin{tabular}{@{}lccc@{}}
\toprule
 & \textbf{Raw MANO} & \textbf{CLUTCH} & \textbf{Ours ($\mathcal{R}_{\text{hand}}$)} \\
\midrule
Rotation repr.        & axis-angle & 6D continuous & 6D continuous          \\
Bimanual root         & per-hand $\mathbf{t}^g$ & per-hand $\mathbf{t}^g$ & wrist midpoint, world $y$ explicit \\
Heading canon.        & \xmark & \xmark & yaw-only, inter-wrist axis \\
Heading repr.         & --     & --     & 6D rotation               \\
Root velocity         & \xmark & \xmark & 2D XZ, heading-aligned    \\
Joint positions       & \xmark & \xmark & \cmark ($32 \times 3$)    \\
Joint velocities      & \xmark & \xmark & \cmark ($32 \times 3$)    \\
Joint rotations       & \cmark ($32 \times 3$) & \cmark ($32 \times 6$) & \cmark ($32 \times 6$) \\
Shape coefficients    & $\boldsymbol{\beta}\in\mathbb{R}^{10}$ & \xmark & \xmark    \\
Total dims (bimanual) & $122$  & $198$  & $\mathbf{393}$            \\
Body-template parity  & SMPL   & SMPL + 6D rot. & extends $\mathcal{R}_{272}$ ($J\!=\!22\!\to\!32$) \\
\bottomrule
\end{tabular}
\end{table}

\subsection{The 201-D essential subset for the TMR evaluator}
\label{app:hand_repr_essential}

The TMR retrieval evaluator (\secref{app:impl_tmr}) does not consume the full $393$-D $\mathcal{R}_{\text{hand}}$ vector. It operates instead on a $201$-dimensional \emph{essential} subset $\mathbf{h}^{\text{ess}}_t$ obtained by dropping the joint-position and joint-velocity blocks of $\mathbf{h}_t$ and keeping the root frame together with the per-joint $6$D rotations:
\begin{equation}
  \mathbf{h}^{\text{ess}}_t \;=\; \bigl[\,
    \underbrace{\mathbf{f}_t}_{9,\;\text{root frame}}\,,\;
    \underbrace{\{\phi(\mathbf{R}^j_t)\}_{j=0}^{31}}_{192,\;\text{joint 6D rotations}}
  \,\bigr] \;\in\; \mathbb{R}^{201}.
  \label{eq:hand_repr_essential}
\end{equation}
In the layout of \tabref{tab:hand_393_layout}, $\mathbf{h}^{\text{ess}}_t$ is the dim union $[0{:}9)\,\cup\,[201{:}393)$ of $\mathbf{h}_t$; the dropped block spans $[9{:}201)$ and contains the $96$ joint positions and the $96$ joint velocities. The same type-aware normalization of \secref{app:hand_repr_normalization} is applied to the surviving $201$ dimensions on both branches of the evaluator.

\paragraph{Why drop positions and velocities.}
The essential subset is the minimal channel set that determines the bimanual hand pose under MANO forward kinematics: given the root frame and the $32$ per-joint $6$D rotations, the dropped positions $\{\tilde{\mathbf{p}}^j_t\}$ and their finite differences $\{\dot{\tilde{\mathbf{p}}}^j_t\}$ are recoverable exactly under the fixed mean MANO shape used throughout (\secref{sec:hand_repr}). The position and velocity blocks of $\mathbf{h}_t$ are therefore redundant with the rotations as far as identifying a pose is concerned as they exist in $\mathcal{R}_{\text{hand}}$ as an inductive bias for continuous generative architectures, exactly as their analogues do in HumanML3D-263 and -272 for body motion, but they do not carry information that the rotations do not.

The redundancy matters at evaluation time because of quantization. VQ-style generators (MoMask's residual VQ, MotionMillion's FSQ) approximate the $\mathbb{R}^{96}$ position and velocity channels through a finite codebook and so reconstruct them less faithfully than the continuous diffusion baselines (Kimodo M, MDM), even when their rotational reconstructions are equally good. An evaluator with access to the position and velocity channels would therefore learn to score motions partly on whether those continuous channels match a continuous reference, systematically penalising VQ generators on a non-articulatory axis. Restricting the TMR evaluator to the rotational $201$-D subset removes that confound: every generator is scored on the channels in which it is on equal footing, and the resulting R@$k$ and FID numbers measure articulation fidelity rather than a mixed signal that conflates articulation with continuous-space reconstruction.

\section{Evaluation Metric Details}
\label{app:metric_details}

Text-conditioned hand motion generation has no per-clip ground truth: many distinct motions can plausibly satisfy the same caption, so a generated clip cannot be scored by direct reference comparison the way one would score a reconstruction or denoising model. As a consequence, no single metric and in particular no single \emph{physical} metric is sufficient to declare a motion ``physically plausible.'' Following the body-motion literature, we therefore triangulate with a bank of complementary measures that fall into two families.

\textbf{Semantic metrics} (\secref{app:semantic_metrics}: FID~\citep{fid}, R-Precision@$k$~\citep{humanml3d}, Matching Score, Diversity, MultiModality) operate in the latent space of the TMR~\citep{tmr} text--motion evaluator (\secref{app:impl_tmr}, trained on the $201$-D essential subset of $\mathcal{R}_{\text{hand}}$, \secref{app:hand_repr_essential}). They jointly measure (i) distributional alignment between generated and GT motions (FID), (ii) text-conditional alignment between a motion and its prompt (R-Precision@$k$, Matching Score), and (iii) within- and cross-prompt variability (Diversity, MultiModality), so a model cannot win all five simultaneously by collapsing to a single mode or by ignoring the caption. Because every value is evaluator-specific, absolute numbers are not directly comparable to HumanML3D body-motion numbers; the reference ranges we cite from body-motion literature should be read as orientation, not as targets to clear. Throughout, $E_m$ and $E_t$ denote the TMR motion and text encoders.

\textbf{Physical metrics} (\secref{app:physical_metrics}) are evaluator-free proxies computed directly on per-joint positions and posed MANO meshes. We borrow smoothness measures (Acceleration, Intra-hand Jerk, HF Power Ratio) and contact-validity measures (Inter- and Self-hand penetration variants) from the body-motion literature~\citep{humanml3d,bimart}, and add two hand-specific axes that we found necessary in pilots: bone-length violation rate, which catches token-by-token drift in autoregressive generators, and a finger-flexion posture mean, which catches the ``flat hand'' failure mode. None of these individually proves a motion is physically valid; together they flag the dominant failure modes we observed in pilots.

\subsection{Semantic Metrics}
\label{app:semantic_metrics}

The arrow next to each metric ($\downarrow$/$\uparrow$) indicates whether smaller or larger is better; metrics whose target is to \emph{match} the GT row of the same table rather than minimise or maximise are flagged inline.

\paragraph{FID~\citep{fid}~\textnormal{($\downarrow$, lower is better).}}
Squared $2$-Wasserstein distance between Gaussians fit to motion-encoder features of the real (GT) and generated test sets:
\[ \text{FID} \;=\; \|\mu - \hat\mu\|_2^2 \;+\; \mathrm{Tr}\!\bigl(\Sigma + \hat\Sigma - 2(\Sigma\,\hat\Sigma)^{1/2}\bigr), \quad \downarrow, \]
where $(\mu, \Sigma)$ is the empirical mean and covariance of $\{E_m(m_i)\}_{i=1}^{N}$ over $N$ GT motions, $(\hat\mu, \hat\Sigma)$ is the same for $\{E_m(\hat{m}_j)\}_{j=1}^{M}$ over $M$ generated motions, $\|\cdot\|_2$ is the Euclidean norm, and $\mathrm{Tr}(\cdot)$ is the matrix trace. For orientation, body-motion generators reported on HumanML3D typically land at $0.05$--$1.0$ for SOTA and $\le 0.05$ on the GT row.

\paragraph{R-Precision@$k$~\citep{humanml3d}~\textnormal{($\uparrow$, higher is better).}}
Let $\mathcal{T} = \{\hat m\}$ denote the set of generated motions over the test split. For each $\hat m \in \mathcal{T}$, the ground-truth caption $t^{\star}$ is placed in a pool of $32$ ($1$ correct $+$ $31$ random distractors drawn uniformly from the test split's captions); the $32$ candidates $\{t_i\}_{i=1}^{32}$ are ranked by ascending cosine distance $d_i = 1 - \cos(E_m(\hat m), E_t(t_i))$ to the generated motion, and a hit is counted when $t^\star$ falls in the top $k$:
\[ \text{R-Precision@}k \;=\; \tfrac{1}{|\mathcal{T}|}\sum_{\hat m \in \mathcal{T}} \mathbb{1}\!\bigl[\,\mathrm{rank}(t^{\star}) \le k\,\bigr] \;\in\; [0, 1], \quad \uparrow, \]
where $\mathbb{1}[\,\cdot\,]$ is the \emph{indicator function} ($1$ if the condition holds, $0$ otherwise), $|\mathcal{T}|$ is the size of the test-generation set, and $\mathrm{rank}(t^{\star})$ is the position of $t^{\star}$ in the ascending-distance ordering of the $32$-candidate pool. Main paper reports $k\!\in\!\{1,3,10\}$; supplemental tables add $k\!\in\!\{2,5\}$. Chance under random retrieval from the pool of $32$ is $k/32$, i.e.\ $0.031, 0.094, 0.313$ at $k\!=\!1,3,10$. On HumanML3D the same protocol typically lands SOTA body-motion generators at $\sim\!0.50$--$0.55$ for $\text{R@}1$ and $\sim\!0.85$--$0.92$ for $\text{R@}10$, with the GT row near the same numbers.

\paragraph{Matching Score (MS, a.k.a.\ MM-Dist)~\textnormal{($\downarrow$, lower is better).}}
Mean paired Euclidean ($L_2$) distance between each generated motion and its ground-truth caption in the TMR latent space, following the HumanML3D convention~\citep{humanml3d}. Let $\{(\hat{m}_i, t_i^{\star})\}_{i=1}^{B}$ be the $B = |\mathcal{T}|$ matched (motion, GT-caption) pairs over the test-generation set:
\[ \text{MS} \;=\; \tfrac{1}{B}\sum_{i=1}^{B} \bigl\| E_m(\hat{m}_i) - E_t(t_i^{\star}) \bigr\|_2 \;\ge\; 0, \quad \downarrow, \]
with $0$ corresponding to perfect text--motion alignment. We additionally report a scale-invariant variant $\text{MS}_{\text{norm}}$ (\tabref{tab:gt_caption_breakdown_app}) that $L_2$-normalizes each embedding to unit length before the distance, bounded in $[0,2]$ and satisfying $\text{MS}_{\text{norm}}^2 = 2 - 2\cos(E_m(\hat{m}), E_t(t^{\star}))$. Because both variants are reported in evaluator-specific units, absolute values do not transfer across evaluators (in particular, HumanML3D body-motion numbers are not directly comparable); we report MS only against our own GT row.

\paragraph{CLIP Score~\textnormal{($\uparrow$, higher is better).}}
Mean paired cosine similarity between each generated motion and its ground-truth caption in the TMR text--motion joint embedding space, the scale-invariant analogue of MS:
\[ \text{CLIP score} \;=\; \tfrac{1}{B}\sum_{i=1}^{B} \cos\!\bigl(E_m(\hat{m}_i),\, E_t(t_i^{\star})\bigr) \;\in\; [-1, 1], \quad \uparrow, \]
with $1$ corresponding to perfect text--motion alignment. We name it ``CLIP score'' by analogy with the image--text CLIP-score convention, although the joint space here is TMR rather than CLIP; this is the headline alignment axis used in the per-category radar plot of \figref{fig:taxonomy_radar}.

\paragraph{Diversity~\textnormal{(target: match GT).}}
Mean Euclidean distance between two random subsets of $S_d = 300$ TMR motion features. Let $\{v_i\}_{i=1}^{S_d}$ and $\{v'_i\}_{i=1}^{S_d}$ be two disjoint random subsets of $\{E_m(\hat m) : \hat m \in \mathcal{T}\}$, each of size $S_d$:
\[ \text{Diversity} \;=\; \tfrac{1}{S_d}\sum_{i=1}^{S_d} \|v_i - v'_i\|_2, \quad \text{target: match GT}. \]
Values substantially below the GT row of the same table indicate mode collapse; values substantially above indicate noise. Body-motion literature reports Diversity in the same units as its evaluator-specific MS, so direct numerical comparison is again only meaningful within a single evaluator.

\paragraph{MultiModality~\textnormal{(target: match GT).}}
Within-prompt analogue of Diversity over $C = 30$ shared prompts with $S_l = 10$ pairs per prompt. For each prompt $c \in \{1, \dots, C\}$ we draw $S_l$ pairs of generations conditioned on the same caption and let $v_{c,i}, v'_{c,i} = E_m(\hat m_{c,i}), E_m(\hat m'_{c,i})$ denote the TMR motion-encoder features of the $i$-th pair:
\[ \text{MModality} \;=\; \tfrac{1}{C \cdot S_l}\sum_{c=1}^{C}\sum_{i=1}^{S_l} \|v_{c,i} - v'_{c,i}\|_2, \quad \text{target: match GT}. \]
Higher indicates more variation under a fixed prompt; excessively high values indicate noise rather than valid alternative interpretations. As with Diversity, the relevant comparison is to the same evaluator's GT row.

All semantic metrics are averaged over 20 independent evaluation runs with 95\% confidence intervals.

\subsection{Physical Metrics}
\label{app:physical_metrics}

Physical metrics are evaluator-free and computed at 30\,fps directly from the per-joint positions obtained by decoding $\mathcal{R}_{\text{hand}}$ and posing the left and right MANO templates by forward kinematics. A single shared routine takes a $(T, 32, 3)$ position array in metres and produces every number reported in the physical half of \tabref{tab:results_main} of the main paper and \tabref{tab:physical_compact_ci} here, for both generator outputs and the GT rows for RoMo Hands, GRAB, and GigaHands. We split the metrics into two families with very different compute profiles.

\emph{Joint-based metrics} (Acceleration, Intra-hand Jerk, HF Power Ratio, Bone-length Violations, Finger Flexion Mean) read the $(T, 32, 3)$ position array directly. They are cheap and we evaluate them over the full RoMo Hands test split.

\emph{Mesh-based metrics} (Inter-hand and Self-hand penetration variants) require posing both $\sim\!778$-vertex MANO meshes every frame and querying the signed point-to-mesh distance from each query (vertex or joint) to a posed mesh, where the sign denotes whether the query lies outside ($d > 0$) or inside ($d < 0$) the mesh; to make the inside/outside test well-defined, each MANO mesh is closed at the wrist with a small triangular cap before the distance query. These mesh queries are roughly two orders of magnitude more expensive than joint-array reads, so we estimate the mesh-based metrics on a randomly sampled 256-sequence subset of the test split at frame stride~4. The reported numbers are therefore order-of-magnitude estimates rather than full-split averages, but the subset is large enough for the rankings to be stable across the 20 evaluation replications used for the 95\% CIs.

Let $\mathbf{x}(t) \in \mathbb{R}^{J \times 3}$ denote joint positions at frame $t \in \{1, \dots, T\}$ ($J = 32$, $16$ per hand; $T$ is the per-clip frame count, with hand $h \in \{L, R\}$ subscripts written as $\mathbf{x}^{h}_{j}(t) \in \mathbb{R}^3$ when needed), $\mathbf{V}^{L}(t), \mathbf{V}^{R}(t)$ the left/right mesh vertices at frame $t$, and $\Delta t = 1/30$\,s the inter-frame interval. Derivatives use central differences:
\begin{align*}
  \dot{\mathbf{x}}(t) &= \frac{\mathbf{x}(t+1) - \mathbf{x}(t-1)}{2\Delta t}, \quad
  \ddot{\mathbf{x}}(t) = \frac{\mathbf{x}(t+1) - 2\mathbf{x}(t) + \mathbf{x}(t-1)}{\Delta t^2}, \\
  \dddot{\mathbf{x}}(t) &= \frac{\ddot{\mathbf{x}}(t+1) - \ddot{\mathbf{x}}(t-1)}{2\Delta t}.
\end{align*}
The physical half of \tabref{tab:results_main} in the main paper is the headline physical-metrics block; it reports Acceleration, Intra-hand Jerk, Inter-hand Penetration Ratio, Self-hand Penetration MAX, and HF Power Ratio. Supplementary axes such as mean inter-hand penetration depth, the BimArt $>\!10$\,mm self-penetration ratio, bone-length violations, and finger flexion mean live in \tabref{tab:physical_compact_ci} here.

\paragraph{Acceleration~\textnormal{(target: match GT).}}
Mean Euclidean acceleration magnitude over all frames and joints:
\[ \text{Acc} \;=\; \tfrac{1}{TJ}\sum_{t=1}^{T}\sum_{j=0}^{J-1} \|\ddot{\mathbf{x}}_{t,j}\|_2 \quad (\text{m/s}^2), \quad \text{target: match GT}, \]
where $\ddot{\mathbf{x}}_{t,j} \in \mathbb{R}^3$ is the central-difference acceleration of joint $j$ at frame $t$ defined above. The target is to \emph{match} the GT row of the same table, not to minimise: under-acceleration indicates over-smoothing and over-acceleration indicates jitter, so deviation in either direction is a quality flag.

\paragraph{Intra-hand Jerk~\textnormal{(target: match GT).}}
Mean Euclidean jerk magnitude after rotating each hand into a per-frame local wrist frame, isolating finger-articulation smoothness from gross arm sweep. For each hand $h \in \{L, R\}$ we build a right-handed orthonormal wrist frame $R^{h}(t) \in \mathrm{SO}(3)$ via Gram--Schmidt on the index-MCP and pinky-MCP joints relative to the wrist root $\mathbf{w}^{h}(t) \in \mathbb{R}^3$ (forward axis: wrist$\to$index-MCP; transverse: index-MCP/pinky-MCP), express each joint locally as
\[ \mathbf{x}^{h,\text{loc}}_{j}(t) \;=\; R^{h}(t)^{\top}\bigl(\mathbf{x}^{h}_{j}(t) - \mathbf{w}^{h}(t)\bigr) \;\in\; \mathbb{R}^3, \]
and apply the same central-difference triple-derivative as above to obtain $\dddot{\mathbf{x}}^{\text{loc}}_{t,j}$ for the joint $j$ in its own hand's local frame:
\[ \text{IntraJerk} \;=\; \tfrac{1}{TJ}\sum_{t=1}^{T}\sum_{j=0}^{J-1} \|\dddot{\mathbf{x}}^{\text{loc}}_{t,j}\|_2 \quad (\text{m/s}^3; \text{ target: match GT}). \]
The wrist derotation removes the gross arm-sweep contributions that otherwise dominate the third derivative on this wide-workspace bimanual dataset (see World-space Jerk below).

\paragraph{Inter-hand Penetration Ratio~\textnormal{($\downarrow$, lower is better).}}
Fraction of frames in which the left and right meshes interpenetrate by at least $1$\,mm:
\[ \text{InterPen} \;=\; \tfrac{1}{T}\sum_{t=1}^{T} \mathbb{1}\!\bigl[\max_{v \in \mathbf{V}^{L}(t)}(-d^{R}_{t}(v)) \,\ge\, 1\,\text{mm} \;\vee\; \max_{v \in \mathbf{V}^{R}(t)}(-d^{L}_{t}(v)) \,\ge\, 1\,\text{mm}\bigr] \quad (\%, \downarrow), \]
where $d^{h}_{t}(v)$ is the signed distance from query vertex $v$ to the hand-$h$ mesh at frame $t$ (positive outside, negative inside; $h \in \{L, R\}$), $\mathbf{V}^{h}(t)$ is the vertex set of mesh $h$ at frame $t$, $\vee$ is logical OR, and $\mathbb{1}[\,\cdot\,]$ is the indicator function defined in the semantic-metric block above. The wrist-cap closure adds a small ($\sim$0.5--1\%) baseline that affects all rows uniformly.

\paragraph{Self-hand Penetration MAX~\textnormal{($\downarrow$, lower is better).}}
Per-sample maximum signed-distance penetration of any joint into its own hand's mesh, averaged across the $S$ test motions. With $\mathbf{x}^{h}_{j}(t)$ as defined above and $d^{h}_{t,\text{rest}}(\,\cdot\,)$ the signed distance to the mesh of hand $h$ at frame $t$, computed with the local rest neighbourhood of the queried joint $j$ excluded so a joint cannot be flagged as penetrating its own bone:
\[ \text{SelfPen}_{\max} \;=\; \tfrac{1}{S}\sum_{s=1}^{S} \max_{t, h, j}\, \max\!\bigl(0,\, -d^{h}_{t,\text{rest}}(\mathbf{x}^{h}_{j}(t))\bigr) \quad (\text{mm}, \downarrow), \]
where the inner $\max(0, \cdot)$ keeps only frames/joints that lie inside the mesh ($-d > 0$). This depth-valued variant replaces the BimArt~\citep{bimart} ``Pen 1\,cm'' frame-ratio in the headline \tabref{tab:results_main} because every FK-correct generator collapses to $0\%$ on the ratio with overlapping CIs, while the depth varies continuously; the BimArt ratio is retained as a supplementary axis in \tabref{tab:physical_compact_ci}.

\paragraph{HF Power Ratio~\textnormal{($\downarrow$, lower is better).}}
Per-joint position power above $f_c = 8$\,Hz, normalised by total power and averaged over joints. Let $\hat{\mathbf{x}}_j(f) \in \mathbb{C}^3$ be the temporal Fourier transform of $\mathbf{x}_j(\cdot)$ over the clip's $T$ frames, evaluated at frequency $f$ in the discrete spectrum $\{0,\, 1/(T \Delta t),\, \dots,\, 1/(2 \Delta t)\}$ (Nyquist $1/(2 \Delta t) = 15$\,Hz at $30$\,fps), and $|\hat{\mathbf{x}}_j(f)|^2 = \sum_{d=1}^{3} |\hat{x}^{(d)}_j(f)|^2$ the squared magnitude summed across the three spatial channels:
\[ \text{HFPR} \;=\; \tfrac{1}{J}\sum_{j=0}^{J-1} \frac{\sum_{f \geq f_c} |\hat{\mathbf{x}}_j(f)|^2}{\sum_{f} |\hat{\mathbf{x}}_j(f)|^2} \;\in\; [0, 1], \quad \downarrow. \]
Real hand motion has a roughly $1/f$ spectrum dropping sharply above $\sim$5\,Hz; excess power above $8$\,Hz signals noise or token-by-token decoding artefacts. HFPR complements Acceleration and Intra-hand Jerk by being normalised for overall motion energy.

\paragraph{Supplementary axes~\textnormal{(\tabref{tab:physical_compact_ci}).}}
\tabref{tab:physical_compact_ci} reports four supplementary physical axes that complement the headline numbers from \tabref{tab:results_main} without replacing them.
\emph{Inter-hand Penetration Depth} (mm; $\downarrow$) is the mean over interpenetrating frames of the maximum left/right interpenetration depth which is a continuous-valued complement to the binary headline Inter-hand Penetration Ratio.
\emph{Self-hand Penetration Ratio} ($\%$ of frames; $\downarrow$) is the original BimArt~\citep{bimart} ``Pen 1\,cm'' convention (fraction of frames where any joint penetrates its own-hand mesh by $>\!10$\,mm), retained here for cross-paper comparability although it discriminates poorly on RoMo Hands.
\emph{Bone-length Violations} ($\%$ of frames; $\downarrow$) is the fraction of frames where any bone deviates from its MANO template length by more than a per-joint threshold ($50/35/30$\,mm at root/PIP/DIP, calibrated from the empirical RoMo Hands maximum); FK-correct generators score $0\%$ by construction, while discrete-token models drift token-by-token.
\emph{Finger Flexion Mean} (m; target: GT) is the time-and-finger mean of fingertip-to-MCP distance, a coarse posture proxy that catches the ``flat hand'' failure mode.

\paragraph{Reference rows.}
Both physical tables include italicised GT rows for RoMo Hands, GRAB~\citep{grab}, and GigaHands~\citep{gigahands}; the latter two are streamed through the same $\mathcal{R}_{\text{hand}}$ decoder and scoring routine as our generators. GRAB (slow sit-and-grasp on a small workspace) provides a clean lower bound for finger-articulation smoothness, while GigaHands is closer to RoMo Hands in workspace and variety. Reference rows are excluded from per-column ``best'' bolding among generators and serve only to anchor the absolute scale of the physical metrics, which are unitful and otherwise hard to interpret in isolation.

\section{Full Experimental Results}
\label{app:results_full}

\tabref{tab:semantic_full} extends the semantic half of \tabref{tab:results_main} with the full retrieval grid (R@1, R@2, R@3, R@5, R@10) plus MM-Dist, FID, and Diversity, all with 95\% CIs. \tabref{tab:physical_compact_ci} adds the supplementary physical axes (inter-hand penetration depth, BimArt $>\!10$\,mm self-penetration ratio, bone-length violations, finger flexion mean) that are useful for cross-checking the headline numbers and for cross-paper comparability. The per-caption-level GT breakdown referenced in \secref{sec:results_main} follows. Unless noted, all semantic evaluations use the TMR evaluator trained with type-aware normalization and the SnapMoGen pool-of-32 protocol with 20 replications.

\begin{table}[h]
\centering
\caption{\textbf{Full semantic scorecard} (all R@$k$, MM-Dist, FID, Diversity) for the generators we benchmark in the paper body. Means $\pm$95\% CI. MotionMillion = v6, CFG $5.0$ (matches Table~\ref{tab:results_main}). Diversity is reported with ``$\rightarrow$'' to indicate that closer to the ground-truth value is better.}
\label{tab:semantic_full}
\resizebox{\textwidth}{!}{%
\begin{tabular}{lcccccccc}
\toprule
Method & R@1 $\uparrow$ & R@2 $\uparrow$ & R@3 $\uparrow$ & R@5 $\uparrow$ & R@10 $\uparrow$ & MM-Dist $\downarrow$ & FID $\downarrow$ & Diversity $\rightarrow$ \\
\midrule
Ground Truth & $63.77 \pm 0.34$ & $76.25 \pm 0.31$ & $82.59 \pm 0.29$ & $89.49 \pm 0.23$ & $96.13 \pm 0.11$ & $69.04 \pm 0.02$ & $0.000 \pm 0.000$ & $120.41 \pm 0.41$ \\
\midrule
MDM (DiP)~\citep{mdm} & $42.28 \pm 0.21$ & $57.30 \pm 0.42$ & $66.15 \pm 0.43$ & $76.95 \pm 0.28$ & $88.92 \pm 0.15$ & $80.19 \pm 0.02$ & $0.144 \pm 0.000$ & $113.44 \pm 0.40$ \\
MoMask~\citep{momask} & $\mathbf{54.00 \pm 0.41}$ & $\mathbf{70.68 \pm 0.36}$ & $\mathbf{78.62 \pm 0.24}$ & $\mathbf{86.07 \pm 0.24}$ & $\mathbf{93.88 \pm 0.14}$ & $74.52 \pm 0.02$ & $0.129 \pm 0.000$ & $115.27 \pm 0.34$ \\
Kimodo M~\citep{kimodo} & $53.82 \pm 0.37$ & $69.04 \pm 0.29$ & $76.99 \pm 0.28$ & $85.56 \pm 0.16$ & $93.76 \pm 0.13$ & $\mathbf{73.67 \pm 0.02}$ & $\mathbf{0.087 \pm 0.000}$ & $\mathbf{116.34 \pm 0.34}$ \\
MotionMillion~\citep{motionmillion} & $31.26 \pm 0.46$ & $43.97 \pm 0.46$ & $51.55 \pm 0.43$ & $63.40 \pm 0.42$ & $82.37 \pm 0.46$ & $87.10 \pm 0.000$ & $0.314 \pm 0.000$ & $109.14 \pm 0.49$ \\
\bottomrule
\end{tabular}}
\end{table}

\begin{table}[ht]
\centering
\caption{\textbf{Supplementary physical metrics on the RoMo Hands test set.} Complements the physical half of the headline \tabref{tab:results_main} with the mean inter-hand penetration depth, the original BimArt~\citep{bimart} $>\!10$\,mm self-hand penetration frame ratio (retained for cross-paper comparability), bone-length violation rate (token-by-token AR-drift signal), and a finger-flexion posture proxy that catches the ``flat hand'' failure mode. Means $\pm$ 95\% CI over 20 replications on a 256-sequence subset (stride 4).}
\label{tab:physical_compact_ci}
\setlength{\tabcolsep}{3pt}
\footnotesize
\begin{tabular}{lcccc}
\toprule
\textbf{Method}
  & \textbf{Inter-hand Pen Depth}
  & \textbf{Self-hand Pen Ratio}
  & \textbf{Bone-length Viol.}
  & \textbf{Finger Flexion} \\
  & (mean, mm; $\downarrow$)
  & ($\%$ frames $>\!10$\,mm; $\downarrow$)
  & ($\%$ frames; $\downarrow$)
  & (mean, m; target: GT) \\
\midrule
\textit{Ground Truth}                          & \textit{$0.081 \pm 0.052$} & \textit{$0.00 \pm 0.00$}  & \textit{$0.00 \pm 0.00$}  & \textit{$0.1120 \pm 0.0010$} \\
\textit{GRAB GT (ref.)}~\citep{grab}           & \textit{$0.000 \pm 0.000$} & \textit{$0.00 \pm 0.00$}  & \textit{$0.00 \pm 0.00$}  & \textit{$0.1217 \pm 0.0007$} \\
\textit{GigaHands GT (ref.)}~\citep{gigahands} & \textit{$0.000 \pm 0.000$} & \textit{$0.00 \pm 0.00$}  & \textit{$0.00 \pm 0.00$}  & \textit{$0.1251 \pm 0.0005$} \\
\midrule
MDM (DiP)~\citep{mdm}                & $0.243 \pm 0.112$           & $1.05 \pm 0.84$           & $1.93 \pm 0.28$           & $0.1035 \pm 0.0014$ \\
MoMask~\citep{momask}                & $0.190 \pm 0.170$           & $0.07 \pm 0.14$           & $6.34 \pm 0.56$           & $0.1136 \pm 0.0016$ \\
Kimodo M~\citep{kimodo}              & $\mathbf{0.150 \pm 0.093}$  & $\mathbf{0.00 \pm 0.00}$  & $1.54 \pm 0.19$           & $0.1127 \pm 0.0014$ \\
MotionMillion~\citep{motionmillion}  & $0.200 \pm 0.114$           & $\mathbf{0.00 \pm 0.00}$  & $\mathbf{0.08 \pm 0.02}$  & $0.0906 \pm 0.0012$ \\
\bottomrule
\end{tabular}
\end{table}

\begin{table}[ht]
\centering
\caption{\textbf{Human ratings support caption fidelity across all five detail levels.} Twelve raters produced 370 ratings over 250 randomly sampled validation clips. Questions (i) and (ii) assess overall action match and hand-specific caption--motion faithfulness on a five-point Likert scale; the final two columns report the fraction rated at least 4.}
\label{tab:user_study}
\small
\setlength{\tabcolsep}{6pt}
\begin{tabular}{@{}lrrrrr@{}}
\toprule
\textbf{Caption level} & \textbf{$N$}
  & \textbf{(i) Mean}
  & \textbf{(i) $\geq 4$}
  & \textbf{(ii) Mean}
  & \textbf{(ii) $\geq 4$} \\
\midrule
Tag       & 74  & 4.51 & 89\% & 3.96 & 69\% \\
Short     & 74  & 4.19 & 85\% & 3.81 & 70\% \\
Medium    & 74  & 4.15 & 82\% & 3.76 & 64\% \\
Detailed  & 74  & 4.03 & 78\% & 3.70 & 64\% \\
Paragraph & 74  & 3.92 & 72\% & 3.74 & 65\% \\
\midrule
\textbf{All} & \textbf{370} & \textbf{4.16} & \textbf{81\%} & \textbf{3.79} & \textbf{66\%} \\
\bottomrule
\end{tabular}
\end{table}

\paragraph{Representation-ablation protocol.}
\tabref{tab:representation_ablation_full} reports the generation-side ablation of $\mathcal{R}_{\text{hand}}$ against raw MANO-102 and the CLUTCH-198 representation. The Kimodo~M variants share one sweep protocol, each reported at its own best validation checkpoint. This differs from the original Kimodo~M run behind \tabref{tab:results_main}, hence the higher absolute numbers (R@1 0.665 versus 0.538); comparisons within the table are unaffected, and \tabref{tab:results_main} keeps the original runs so that all four generators are reported under one shared protocol. A further Kimodo~M variant isolates the redundant channels: removing the velocity and heading channels (289-D) lowers R@1 to 0.639, level with the rotation-only baselines, so the velocity and heading channels, rather than the joint positions, carry the improvement.

\begin{table}[ht]
\centering
\caption{\textbf{Full representation-ablation results.} Each model is trained under a representation-matched sweep with all non-representation choices fixed within its block. Diversity should approach the corresponding GT value (MoMask: 120.41; Kimodo M: 120.95). CLIP was not reported for the MoMask sweep. Best within each model block is \textbf{bold}.}
\label{tab:representation_ablation_full}
\small
\setlength{\tabcolsep}{4pt}
\resizebox{\textwidth}{!}{%
\begin{tabular}{@{}llrrrrrrr@{}}
\toprule
\textbf{Generator} & \textbf{Representation}
  & \textbf{R@1} $\uparrow$
  & \textbf{R@3} $\uparrow$
  & \textbf{R@10} $\uparrow$
  & \textbf{FID} $\downarrow$
  & \textbf{Diversity} $\rightarrow$
  & \textbf{MM-Dist} $\downarrow$
  & \textbf{CLIP} $\uparrow$ \\
\midrule
MoMask & $\mathcal{R}_{\text{hand}}$-393 & \textbf{0.547} & \textbf{0.794} & \textbf{0.948} & \textbf{0.153} & \textbf{114.99} & \textbf{0.941} & -- \\
       & MANO-102                         & 0.395          & 0.658          & 0.888          & 0.221          & 111.37          & 1.049          & -- \\
       & CLUTCH-198                       & 0.368          & 0.617          & 0.862          & 0.269          & 109.95          & 1.077          & -- \\
\addlinespace[3pt]
Kimodo M & $\mathcal{R}_{\text{hand}}$-393 & \textbf{0.665} & \textbf{0.871} & 0.971          & 0.114          & 116.09          & \textbf{0.868} & \textbf{0.601} \\
         & MANO-102                         & 0.646          & 0.868          & \textbf{0.980} & 0.135          & 115.15          & 0.896          & 0.581 \\
         & CLUTCH-198                       & 0.646          & 0.856          & 0.974          & \textbf{0.106} & \textbf{117.17} & 0.882          & 0.590 \\
\bottomrule
\end{tabular}}
\end{table}

\begin{table}[ht]
\centering
\caption{\textbf{Removing source-video overlap preserves the benchmark ranking.} Metrics are recomputed after removing every test clip whose source video appears in the training set. Parentheses give the change from the corresponding segment-level result in \tabref{tab:results_main}; R@1 is shown as a fraction here.}
\label{tab:video_disjoint_audit}
\small
\setlength{\tabcolsep}{5pt}
\begin{tabular}{@{}lrrrr@{}}
\toprule
\textbf{Method}
  & \textbf{R@1} $\uparrow$
  & \textbf{FID} $\downarrow$
  & \textbf{MM-Dist} $\downarrow$
  & \textbf{Diversity} $\rightarrow$ \\
\midrule
MoMask        & 0.549 {\scriptsize $(+0.009)$} & 0.158 {\scriptsize $(+0.029)$} & 0.922 {\scriptsize $(-0.023)$} & 113.9 {\scriptsize $(-1.4)$} \\
Kimodo M      & 0.545 {\scriptsize $(+0.007)$} & 0.116 {\scriptsize $(+0.029)$} & 0.920 {\scriptsize $(-0.022)$} & 114.2 {\scriptsize $(-2.1)$} \\
MDM (DiP)     & 0.405 {\scriptsize $(-0.018)$} & 0.171 {\scriptsize $(+0.027)$} & 1.030 {\scriptsize $(-0.001)$} & 114.0 {\scriptsize $(+0.6)$} \\
MotionMillion & 0.287 {\scriptsize $(-0.026)$} & 0.322 {\scriptsize $(+0.008)$} & 1.130 {\scriptsize $(+0.025)$} & 107.7 {\scriptsize $(-1.4)$} \\
\midrule
\textit{Ground Truth} & \textit{0.597 {\scriptsize $(-0.041)$}} & \textit{0.000 {\scriptsize $(+0.000)$}} & \textit{0.880 {\scriptsize $(+0.016)$}} & \textit{119.1 {\scriptsize $(-1.3)$}} \\
\bottomrule
\end{tabular}
\end{table}

\begin{table}[ht]
\centering
\caption{\textbf{Mid-length captions (L2--L3) retrieve best;} the verbose L4 full-paragraph captions hurt slightly versus mid-length descriptions. Levels L0--L4 run from short headline $\rightarrow$ full paragraph; \emph{Random} matches \tabref{tab:results_main} (uniform level draw per replication).}
\label{tab:gt_caption_breakdown_app}
\resizebox{0.95\textwidth}{!}{%
\begin{tabular}{lcccccc}
\toprule
Metric & \textbf{L0} & \textbf{L1} & \textbf{L2} & \textbf{L3} & \textbf{L4} & \textbf{Random (uniform)} \\
\midrule
R@1 $\uparrow$  & $53.02 \pm 0.41$ & $65.04 \pm 0.40$ & $\mathbf{68.43 \pm 0.40}$ & $66.61 \pm 0.38$ & $65.67 \pm 0.26$ & $63.77 \pm 0.34$ \\
R@2 $\uparrow$  & $67.52 \pm 0.34$ & $77.21 \pm 0.32$ & $\mathbf{80.08 \pm 0.28}$ & $79.03 \pm 0.24$ & $78.27 \pm 0.31$ & $76.25 \pm 0.31$ \\
R@3 $\uparrow$  & $75.22 \pm 0.26$ & $83.16 \pm 0.29$ & $\mathbf{85.50 \pm 0.32}$ & $84.75 \pm 0.26$ & $84.23 \pm 0.26$ & $82.59 \pm 0.29$ \\
R@5 $\uparrow$  & $83.69 \pm 0.24$ & $89.78 \pm 0.23$ & $\mathbf{91.57 \pm 0.23}$ & $91.27 \pm 0.26$ & $90.91 \pm 0.20$ & $89.49 \pm 0.23$ \\
R@10 $\uparrow$ & $93.53 \pm 0.18$ & $96.02 \pm 0.15$ & $96.51 \pm 0.10$ & $\mathbf{96.79 \pm 0.10}$ & $96.70 \pm 0.11$ & $96.13 \pm 0.11$ \\
MM-Dist $\downarrow$       & $73.32$ & $68.24$ & $\mathbf{66.67}$ & $67.86$ & $68.72$ & $69.04$ \\
CLIP score $\uparrow$      & $0.544$ & $0.609$ & $\mathbf{0.628}$ & $0.615$ & $0.606$ & $0.600$ \\
MM-Dist$_\text{norm}$ $\downarrow$ & $0.924$ & $0.852$ & $\mathbf{0.831}$ & $0.848$ & $0.860$ & $0.864$ \\
FID $\downarrow$           & $0.000$ & $0.000$ & $0.000$ & $0.000$ & $0.000$ & $0.000$ \\
Diversity     & \multicolumn{6}{c}{$120.41 \pm 0.41$ (level-independent: pooled across all generated motions)} \\
\bottomrule
\end{tabular}}
\end{table}

\paragraph{Reading the caption-level breakdown.}
\tabref{tab:gt_caption_breakdown_app} shows a non-monotone shape across $L_0\!\to\!L_4$: GT retrieval is weakest at $L_0$ (single-action verbs too coarse to separate 31 distractors), peaks at $L_2$--$L_3$ (atomic finger states with relations are the most discriminative granularity), and decreases at $L_4$ where verbose paragraph paraphrases reintroduce non-discriminative filler tokens. The ``Random'' column is the uniform-over-levels mixture used to train the TMR evaluator and to report numbers in the main paper; it sits below the per-level peak by design, conservatively averaging over all five levels and matching the realistic deployment regime where a downstream user does not know which granularity the system was trained against.

\section{Implementation Details}
\label{app:impl}

This appendix expands \secref{sec:impl} with the architectures, optimization, and training schedules of every model in this work. We document \emph{algorithmic} implementation detail (model size, batch size, LR, schedule, duration, design decisions vs.\ published defaults); compute infrastructure is intentionally omitted.

\subsection{Common Setup}
\label{app:impl_common}

\paragraph{Dataset.}
All RoMo Hands models train on the released LitData shards at 30\,fps (an 85\%/10\%/5\% train/val/test split of the 541{,}929 clips), with clips longer than 196 frames (6.5\,s) randomly cropped to 196 frames and clips shorter than 40 frames (1.3\,s) zero-padded to 40, 32 bimanual joints, and 5 paraphrased captions per clip sampled uniformly during training.

\paragraph{Motion representation and normalization.}
Every model consumes the same $\mathcal{R}_{\text{hand}}$ feature (195 \textsc{gaussian} + 194 \textsc{cyclic} + 4 \textsc{constant} = 393 dims) with the type-aware normalization of \secref{app:hand_repr_normalization} applied at the dataloader stage. Cached $(\mu_i, \sigma_i)$ are computed on the train split and frozen across all downstream eval; the TMR evaluator is fit under the same normalization so retrieval-based metrics live in a representation-consistent latent space.

\paragraph{Text encoding.}
A frozen T5-v1.1-base~\citep{t5} encoder runs inline on GPU (bf16 buffer, fp32 params), tokenized to max 256 tokens (180 for MotionMillion, capped to its FSQ token budget). Kimodo M consumes the pooled output (768-D); all other models cross-attend to the full token sequence. For each clip we hold 5 paraphrased captions of escalating verbosity, and at training time sample uniformly across a 5-bucket length schedule (\texttt{max\_lengths}\,=\,[16, 32, 96, 192, 256]); val cycles through buckets round-robin for length balance.

\paragraph{Optimization defaults} (overrides noted per-model).
AdamW~\citep{adamw}, $\beta\!=\!(0.9, 0.99)$, weight decay $0$ for diffusion / tokenizer heads and $0.01$ for LM-style heads (TMR, MotionMillion LLaMA, Kimodo M). bf16-mixed precision; gradient clipping at $0.5$ (MDM, TMR) or $1.0$ (others). EMA $\beta\!=\!0.999$ every 10 steps on diffusion models only. \texttt{torch.compile} enabled by default; disabled on MoMask Stages~1/2 due to recompile thrash from BERT-style mask-ratio sampling. Validation every 5 epochs (every 1 epoch for TMR).

\subsection{Per-model footprint at convergence}
\label{supp:impl_summary}

\tabref{tab:impl_footprint} summarizes the training footprint of every component at validation convergence (bf16-mixed throughout; ``effective batch'' is per-GPU batch $\times$ \#GPUs with no gradient accumulation).
All four generators are sized inside a single capacity decade ($40$M--$150$M trainable text-conditional parameters), so the cross-architecture comparison in \secref{sec:results_main} reflects representational fit rather than raw scale.
MotionMillion is the only entry that required a material change from its published configuration: its $1$B-parameter LLaMA-style decoder would have been an order of magnitude larger than the next-biggest generator and would have confounded the comparison, so we trained the $150$M variant which was the smallest published scale at which the autoregressive recipe still converges to non-trivial accuracy on a $15$K-token FSQ codebook.
\appref{app:impl_scaling} unpacks each per-generator adaptation; detailed architectures and schedules follow.

\begin{table}[ht]
\centering
\caption{Per-model training footprint at validation convergence. Cross-references point to the full per-model subsubsections below.}
\label{tab:impl_footprint}
\small
\resizebox{\textwidth}{!}{%
\begin{tabular}{@{}llrrrl@{}}
\toprule
\textbf{Model} & \textbf{Family} & \textbf{Params} & \textbf{Eff.~batch} & \textbf{LR ($\times 10^{-4}$)} & \textbf{Val~loss / acc.} \\
\midrule
Kimodo M (\S\ref{app:impl_kimodo})           & Diffusion (DDPM/DDIM)        & 80M  & 1024 & $0.20$ & 0.0810 \\
MDM DiP (\S\ref{app:impl_mdm})               & Diffusion (prefix)           & 40M  & 512  & $3.27$ & \textbf{0.0469} \\
MoMask Stage~0 RVQ (\S\ref{app:impl_momask}) & VQ tokenizer ($6{\times}512$) & 10M  & 2048 & $2.83$ & 0.026 (rec.) \\
MoMask Stage~1 mask (\S\ref{app:impl_momask})    & Masked transformer       & 12M  & 1536 & $3.46$ & 3.79 / 25.9\% \\
MoMask Stage~2 res.~(\S\ref{app:impl_momask})    & Residual transformer     & 12M  & 1536 & $3.46$ & 4.50 / 10.4\% \\
MotionMillion FSQ (\S\ref{app:impl_mm})      & FSQ tokenizer (15.4K)        & 12M  & 512  & $2.00$ & 0.0171 (rec.) \\
MotionMillion LLaMA (\S\ref{app:impl_mm})    & Causal LM (RoPE)             & 150M & 1024 & $2.45$ & 7.36 / 6.4\% \\
TMR evaluator (\S\ref{app:impl_tmr})         & Dual encoder (InfoNCE+KL)    & 80M  & 768  & $2.50$ & 6.38 \\
\bottomrule
\end{tabular}}
\end{table}

\subsection{Generative Models}
\label{app:impl_generative}

\subsubsection{Kimodo M (text-to-motion diffusion, paper-M target)}
\label{app:impl_kimodo}

\paragraph{Architecture.} Single-stack transformer denoiser (8 layers, 8 heads, latent 1024, dropout 0.1, max length 512 tokens, 49 learnable conditioning tokens prepended). Text condition is the pooled T5-v1.1-base vector (768-D) projected and concatenated with the conditioning block. Attention is SDPA (PyTorch~2 fused).

\paragraph{Diffusion.} 1000-step training schedule; 100-step DDIM at inference; CFG dropout $0.1$, CFG weight $2.5$. Variance fixed.

\paragraph{Optimization.} AdamW ($\beta\!=\!(0.9, 0.99)$, no decay), LR $2 \times 10^{-5}$ held flat (Kimodo~\citep{kimodo} does not LR-scale with batch). EMA $\beta\!=\!0.995$, update every 10 steps. Effective batch $1024$ ($256\!\times\!4$ GPUs), matching the published ``medium'' config. The registered checkpoint is at epoch 1919 of the 2300-epoch budget (val/loss $0.0810$); val/loss reached its $0.082$ floor $\sim$700 epochs earlier and the tail was within the noise band.

\subsubsection{MDM (Motion Diffusion Model)}
\label{app:impl_mdm}

We adopt the DiP (Diffusion-in-Prefix) variant of MDM~\citep{mdm}, based on the \texttt{trans\_dec} architecture (transformer \emph{decoder} denoiser, conditioning concatenated with the noisy motion sequence as a prefix token) rather than MDM's original encoder-style architecture, which we found underperformed in earlier sweeps.
The model is conditioned on a 20-frame prefix and predicts the next 40 frames rather than generating the full 196-frame clip unconditionally; this collapses most of the multi-modal posterior and lets a much shorter sampling schedule suffice.

\paragraph{Architecture.} 8 layers, 8 attention heads, hidden dim 512, feed-forward 2048, dropout $0.1$, GELU activations, SDPA attention. The text condition is the full T5-v1.1-base token sequence cross-attended (via prefix tokens) into every layer. Per-channel loss weights match the representation: position dims (9--104) get weight 2; root XZ velocity and height (0--2) get weight 10 (under-represented but high-impact for trajectory).

\paragraph{Sequence shape.} Fixed window of 20 frames context $+$ 40 frames prediction $= 60$ total, with both ends pinned (\texttt{min\_seq\_length = max\_seq\_length = 60}).
Inference requires a prefix tensor (in practice: the ground-truth prefix from the eval pipeline).

\paragraph{Diffusion.} Cosine 50-step DDPM training schedule with variance fixed-small, CFG dropout $0.1$, and CFG weight $2.5$ at sampling.
At validation we reduce to 10 DDPM steps: prefix-conditioned generation empirically needs an order of magnitude fewer denoising steps than the full-sequence variant.

\paragraph{Optimization.} AdamW, learning rate $3.27 \times 10^{-4}$ (the batch-size-scaled paper baseline), gradient clip $0.5$, EMA on ($\beta = 0.999$, update every 10 steps).
Per-GPU batch $256$ across 2 GPUs (effective batch $512$). 200 epochs. Best val/loss $= 0.04689$ at epoch 74.

\paragraph{Scale-down rationale.} The $196 \to 60$-frame reduction is a deliberate representation simplification, not a compute compromise: when the model predicts only 40 frames given 20 frames of context, the DDPM posterior is far less multimodal and 10 denoising steps suffice.

\subsubsection{MoMask (residual VQ + masked transformer)}
\label{app:impl_momask}

MoMask~\citep{momask} is a 3-stage pipeline: (Stage~0) train a residual VQ tokenizer that maps motion to a discrete sequence; (Stage~1) train a masked transformer that fills in masked tokens conditional on text; (Stage~2) train a residual transformer that refines the higher-residual codes given the Stage~1 output.
Stage~1 and Stage~2 are independent given a frozen Stage~0 tokenizer and were trained in parallel on independent jobs.

\paragraph{Stage~0: RVQ tokenizer.}
Convolutional encoder ($\text{down\_t}\!=\!2$, stride 2 $\to$ token rate 7.5\,Hz), 6-quantizer residual VQ with 512 codes per quantizer (per-quantizer codebooks, not shared), code dim 512, EMA codebook updates ($\mu\!=\!0.99$), 20\% code dropout. 64-frame training window. Text-independent; AdamW, LR $2.83\times 10^{-4}$, no decay/EMA, step LR at iters $\{150\text{K}, 250\text{K}\}$ with $\gamma\!=\!0.05$. 200 epochs at single-GPU batch $2048$. Reconstruction val/loss $0.026$, codebook perplexity $156$ (moderate utilization typical for first-stage VQs at this dataset scale).

\paragraph{Stage~1: Masked transformer.}
Operates on quantizer~1. At each step a cosine-scheduled fraction of tokens is replaced by \texttt{[MASK]} and the model predicts originals given remaining context and the text caption. \emph{Architecture:} 8 layers, 6 heads, latent 384, FFN 1024, dropout $0.2$, max length 55 tokens. Text is the full T5 sequence cross-attended into every layer; caption-drop $0.1$. \emph{Optimization:} AdamW, LR $3.46\times 10^{-4}$, step at iters $\{72\text{K}, 100\text{K}\}$ with $\gamma\!=\!0.1$, 1K linear warmup; 120 epochs at effective batch $1536$ ($768\!\times\!2$ GPUs). Val/loss $3.79$, accuracy $25.85\%$. Batch was scaled $1.5\!\times$ over the H200 baseline with $\sqrt{1.5}$-scaled LR; dropout was bumped from MoMask's $0.1$ to $0.2$ because the hand dataset is $\sim$5$\times$ smaller than HumanML3D ($+1.5$\,pp val accuracy). \texttt{torch.compile} was disabled after observing recompile thrash from BERT-style mask-ratio sampling.

\paragraph{Stage~2: Residual transformer.}
Operates on quantizers 2--6 conditional on the Stage-1 prediction. Same architecture as Stage~1 except: conditioning input is the Stage-1 token sequence (text still cross-attended), \texttt{cond\_drop\_prob}$\!=\!0.2$, weights shared across the 5 residual heads. \emph{Optimization:} identical schedule to Stage~1; val/loss $4.50$, accuracy $10.39\%$. Stage~2 was still on a slow downward slope at epoch 120 but the schedule was kept identical for parallel-launch simplicity; extending to 200 epochs would likely move accuracy to 13--15\%. Stages~1 and~2 share no weights and were trained in parallel on disjoint 2-GPU shards once the Stage-0 checkpoint was available.

\subsubsection{MotionMillion (FSQ tokenizer + LLaMA AR decoder)}
\label{app:impl_mm}

A two-stage tokenizer + autoregressive language model, modeled after MotionMillion~\citep{motionmillion}.
We trained the small (150M-parameter) variant.

\paragraph{Stage~1: FSQ tokenizer.}
Finite Scalar Quantization~\citep{fsq} replaces MoMask's residual VQ; FSQ partitions a low-dim latent into discrete bins per axis, sidestepping codebook collapse. \emph{Architecture:} the same convolutional encoder as MoMask's Stage~0 ($\text{down\_t}\!=\!2$, token rate 7.5\,Hz, width 512, depth 3, 64-frame window), with 5-axis FSQ at levels $[8,8,8,6,5]$ for an implicit codebook of $15{,}360$ tokens ($\approx\!30\!\times$ MoMask's 512). \emph{Optimization:} AdamW, LR $2\times 10^{-4}$, step at iters $\{150\text{K}, 250\text{K}\}$ with $\gamma\!=\!0.05$; 600 epochs at single-GPU batch $512$. Val/loss $0.0171$ at epoch 569, about $4\times$ lower reconstruction loss than the MoMask Stage-0 RVQ, as expected for a $30\times$ codebook.

\paragraph{Stage~2: LLaMA AR decoder.}
Causal LLaMA-style~\citep{llama} model predicting the next FSQ token given (text, previous tokens). \emph{Architecture:} 12 layers, 12 heads (no GQA), hidden 768, intermediate 2048, RoPE~\citep{rope} $\theta\!=\!10\text{K}$, RMSNorm, SDPA. Max motion-token length 64 ($\approx\!8.5$\,s). Text condition is the full T5 sequence (180-token cap) cross-attended rather than concatenated, keeping the AR prefix short. CFG drop $0.1$, CFG scale $2.5$. \emph{Optimization:} AdamW, LR $2.45\times 10^{-4}$ ($\sqrt{1.5}$-scaled from the H200 baseline), weight decay $0.01$, gradient clip $1.0$, step at iters $\{30\text{K}, 60\text{K}\}$ with $\gamma\!=\!0.1$, 2K warmup; effective batch $1024$ ($128\!\times\!8$ GPUs). \texttt{torch.compile} enabled on the inner LlamaModel only. \emph{Outcome:} best val/loss $7.36$ (accuracy $6.36\%$) at epoch 79 of a 250-epoch budget with early stopping (patience 15 on val/loss); stop fired at epoch 154. Train accuracy at the best checkpoint was $\sim$58\%, indicating the 150M variant overfits this dataset. The original MotionMillion uses a 1B-parameter decoder; we started with 150M to de-risk the FSQ integration, and the train/val gap is the green light for capacity scale-up. 1B and 3B configs are checked in but untrained.

\subsection{Evaluation: TMR retrieval evaluator}
\label{app:impl_tmr}

The TMR model~\citep{tmr} is the \emph{evaluator}, not a generator: once trained it is the canonical scorer for every text-to-motion model in this work via R@$k$ retrieval and FID.

\paragraph{Architecture.} Dual encoder with a shared transformer backbone, consuming the $201$-D essential subset of $\mathcal{R}_{\text{hand}}$ (root frame + per-joint $6$D rotations only; \secref{app:hand_repr_essential}). Both branches encode into a 768-D joint latent space. 8 layers, 8 heads, FFN 3072, dropout $0.1$, SDPA; T5-v1.1-base inline (256-token cap, bf16 buffer).

\paragraph{Loss.} Standard TMR combination: reconstruction ($\lambda_\text{rec}\!=\!1.0$), motion-latent KL ($\lambda_\text{KL}\!=\!10^{-5}$), encoder regularization ($\lambda_E\!=\!10^{-5}$), text--motion latent alignment ($\lambda_\text{latent}\!=\!10^{-5}$), and InfoNCE~\citep{infonce} contrastive across the batch ($\lambda_\text{NCE}\!=\!0.1$, temperature $0.1$). Unlike SnapMoGen~\citep{snapmogen}, we do not use the \texttt{threshold\_selfsim$\,$=$\,$0.80} paraphrase mask: in hyperparameter sweeps, removing the mask and using AdamW weight decay $0.01$ (vs.\ SnapMoGen's $10^{-5}$) gave the best R@1 by $\sim$3--4\,pp on the held-out val set.

\paragraph{Optimization.} AdamW ($\beta\!=\!(0.9,0.99)$), LR $2.5\times 10^{-4}$, weight decay $0.01$, gradient clip $0.5$ (tighter than diffusion models because contrastive loss is sensitive to spikes from hard negatives), 1500-step linear warmup, step LR at epochs $\{40, 60\}$ with $\gamma\!=\!0.1$. 100 epochs at effective batch $768$ ($384\!\times\!2$ GPUs); \texttt{torch.compile} enabled. Validation every epoch (rather than every 5) so the best-by-R@1 checkpoint is reliably captured. Val/loss $6.38$ at epoch 98; the top-6 checkpoints across epochs 93--99 are within $0.004$ val/loss of each other. \emph{All retrieval-based metrics (R@$k$, FID, Diversity, MModality) in this paper are computed against this evaluator.}

\subsection{Matched-budget scaling across architectures}
\label{app:impl_scaling}

The cross-architecture comparison in \secref{sec:results_main} is a comparison of \emph{representational fit} between four generator families on the same data, not a comparison of compute. To keep that comparison fair we sized every generator under two principles: (i)~\emph{matched to the data}: depth, batch, and schedule are tuned to RoMo Hands' 85\% train split rather than carried over from each method's body-motion default; and (ii)~\emph{matched to one another}: generator parameter counts are kept inside a single decade ($\sim\!40$M--$150$M) so any ranking we report reflects how each architecture handles the bimanual representation, not raw capacity. Tokenizers (MoMask RVQ, MotionMillion FSQ) are exempt from~(ii) because their objective is reconstruction quality and codebook coverage, not text-conditional generation; we keep them at their published sizes and let the codebook do the work.

The headline consequence is the MotionMillion entry: its as-published $1$B-parameter decoder would have been $\sim\!7\!\times$ larger than the next biggest generator and would have confounded the cross-architecture comparison, so we trained the $150$M variant in this work instead. \tabref{tab:impl_scaling} summarizes the resulting band.

\begin{table}[ht]
\centering
\caption{\textbf{Generator capacities are matched to RoMo Hands' size and to one another within a single decade ($\sim\!40$M--$150$M).} ``Generator params'' counts only the trainable text-conditional model; tokenizers (RVQ, FSQ) are excluded because they are not part of the cross-architecture comparison. The only material capacity change versus the published configurations is MotionMillion's $1\text{B}\!\to\!150\text{M}$ scale-down.}
\label{tab:impl_scaling}
\small
\begin{tabular}{@{}lrrp{0.42\linewidth}@{}}
\toprule
\textbf{Generator} & \textbf{Generator} & \textbf{Published} & \textbf{Adaptation to RoMo Hands} \\
                   & \textbf{params}    & \textbf{params}    &                                  \\
\midrule
Kimodo M           & 80M  & 80M           & Paper ``medium'' config held; DDPM/DDIM schedule unchanged. \\
MDM (DiP)          & 40M  & 40M           & Sequence shortened ($196\!\to\!60$), DDPM steps $50\!\to\!10$; prefix conditioning collapses the posterior. \\
MoMask Stage~1     & 12M  & 12M           & Batch $\!\times\!1.5$ with $\sqrt{1.5}$-scaled LR; dropout $0.1\!\to\!0.2$ for the smaller dataset. \\
MoMask Stage~2     & 12M  & 12M           & Same schedule as Stage~1; weights shared across the 5 residual heads. \\
MotionMillion      & 150M & \textbf{1B}   & Decoder shrunk $\sim\!7\!\times$ to keep the comparison inside the same parameter band. \\
\midrule
TMR (evaluator)    & 80M  & 80M           & Paraphrase mask removed; weight decay $0.01$ ($+$3--4\,pp R@1 on val). \\
\bottomrule
\end{tabular}
\end{table}

\paragraph{Two clarifications.}
First, MoMask is reported as two rows because Stages~1 and~2 are independent transformers given the frozen Stage~0 tokenizer; they share the architecture and the schedule. The Stage~0 RVQ (10M params, $6\!\times\!512$ codes) and the MotionMillion FSQ (12M params, $15.4$K-token codebook) are kept at their published sizes for the reasons given above. Second, the MotionMillion train/val accuracy gap reported in \secref{app:impl_mm} ($\sim\!58\%$/$6\%$) is the green light for revisiting the $1$B and $3$B configurations once the rest of the pipeline is locked, but doing so under a fair-comparison protocol would also require pushing the other generators' capacities up by a comparable factor; we leave that scale-up to future work.

\subsection{Reproducibility}
\label{app:impl_repro}

The RoMo Hands dataset, the action taxonomy, the five-tier captions, the TMR retrieval evaluator, and the per-baseline training configurations and checkpoints reported in this paper will be publicly released through the project page under the licenses listed in \secref{app:licenses}. Together with the architectures, optimizers, and training budgets documented in \secref{app:impl}, this release will allow any single result in \tabref{tab:results_main} to be reproduced from the same starting point.

\section{Licenses for External Assets}
\label{app:licenses}

The third-party assets used in this work are listed below. Each is cited in the main paper and used under its published license.

\begin{itemize}\denselist
    \item \textbf{RoMo}~\citep{romo_body}: source video dataset and the curation pipeline we extend. Non-commercial research.
    \item \textbf{WiLoR}~\citep{wilor}: monocular bimanual hand detector and pose estimator used in the same pipeline. Non-commercial research.
    \item \textbf{MANO}~\citep{mano}: parametric hand model that defines the joint hierarchy and blendshapes underlying our motion representation. MPI non-commercial research license.
    \item \textbf{SMPL-X}~\citep{smplx}: full-body parametric human model that we use to source the SMPL-X-aligned wrist trajectories in the wrist-anchoring step. MPI non-commercial research license.
    \item \textbf{Qwen3.5-27B}~\citep{qwen3.5}: VLM used to generate the five-level captions. Apache License 2.0 accompanying the released checkpoint.
    \item \textbf{T5-v1.1-base}~\citep{t5}: text encoder inside the TMR retrieval evaluator. Apache License 2.0.
    \item \textbf{MDM}~\citep{mdm}: \texttt{motion-diffusion-model} reference implementation, used with the DiP variant adaptations. MIT License.
    \item \textbf{MoMask}~\citep{momask}: \texttt{momask-codes} reference implementation, used with our per-baseline adaptations. MIT License.
    \item \textbf{MotionMillion}~\citep{motionmillion}: \texttt{MotionMillion-Codes} reference implementation by Shanghai AI Laboratory, used with the $150$M-variant adaptations. Apache License 2.0. (Note: the released MotionMillion \emph{dataset} is CC BY-NC-SA 4.0, but we do not use it; we train on RoMo Hands.)
    \item \textbf{Kimodo}~\citep{kimodo}: NVIDIA's \texttt{nv-tlabs/kimodo} reference implementation, used with the $\mathcal{R}_{\text{hand}}$ I/O dimension change. Apache License 2.0 for the codebase; the separately licensed pretrained checkpoints (NVIDIA Open Model License) are not used, since we train from scratch on RoMo Hands.
\end{itemize}

\end{document}